\documentclass[letterpaper]{article} 
\usepackage{aaai2026}  
\usepackage{times}  
\usepackage{helvet}  
\usepackage{courier}  
\usepackage[hyphens]{url}  
\usepackage{graphicx} 
\usepackage{natbib}  
\usepackage{caption} 
\usepackage{algorithm}
\usepackage{algorithmic}

\usepackage{amsmath,amsfonts}
\usepackage{amssymb}
\usepackage{amsthm}

\usepackage{multirow}
\usepackage{makecell}
\usepackage{placeins}  
\usepackage{subcaption}

\usepackage{booktabs}
\nocopyright  
\usepackage{newfloat}
\usepackage{listings}
\DeclareCaptionStyle{ruled}{labelfont=normalfont,labelsep=colon,strut=off} 
\floatstyle{ruled}
\newfloat{listing}{tb}{lst}{}
\floatname{listing}{Listing}
\title{Codebook-Centric Deep Hashing: End-to-End Joint Learning of Semantic Hash Centers and Neural Hash Function}
\author{
    Shuo Yin\textsuperscript{\rm 1},
    Zhiyuan Yin\textsuperscript{\rm 1},
    Yuqing Hou\textsuperscript{\rm 2},
    Rui Liu\textsuperscript{\rm 3},
    Yong Chen\textsuperscript{\rm 1}\footnote{\makebox[0pt][l]{Corresponding author (alphawolf.chen@gmail.com).}},
    Dell Zhang\textsuperscript{\rm 4}
}
\affiliations{
    \textsuperscript{\rm 1}School of Computer Science, Beijing University of Posts and Telecommunications, Beijing, China\\
    \textsuperscript{\rm 2}Independent Researcher\\
    \textsuperscript{\rm 3}School of Computer Science and Engineering, Beihang University, Beijing, China\\
    \textsuperscript{\rm 4}Institute of Artificial Intelligence (TeleAI), China Telecom, Shanghai, China\\
     yinshuo06@163.com, zy\_yin@bupt.edu.cn, houyuqing1988@gmail.com, lr@buaa.edu.cn, dell.z@ieee.org
    
}

\usepackage{bibentry}

\begin{document}

\maketitle
\insert\footins{\noindent\footnotesize This paper has been accepted for publication at the 40th Annual
 AAAI Conference on Artificial Intelligence (AAAI-26).\par}

\begin{abstract}
Hash center-based deep hashing methods improve upon pairwise or triplet-based approaches by assigning fixed hash centers to each class as learning targets, thereby avoiding the inefficiency of local similarity optimization. However, random center initialization often disregards inter-class semantic relationships. While existing two-stage methods mitigate this by first refining hash centers with semantics and then training the hash function, they introduce additional complexity, computational overhead, and suboptimal performance due to stage-wise discrepancies.
To address these limitations, we propose \textbf{Center-Reassigned Hashing (CRH)}, an end-to-end framework that \textbf{dynamically reassigns hash centers} from a preset codebook while jointly optimizing the hash function. Unlike previous methods, CRH adapts hash centers to the data distribution \textbf{without explicit center optimization phases}, enabling seamless integration of semantic relationships into the learning process. Furthermore, \textbf{a multi-head mechanism} enhances the representational capacity of hash centers, capturing richer semantic structures. Extensive experiments on three benchmarks demonstrate that CRH learns semantically meaningful hash centers and outperforms state-of-the-art deep hashing methods in retrieval tasks. 
\end{abstract}

\begin{links}
   \link{Code Project}{https://github.com/iFamilyi/CRH}
   \link{Conference Version}{https://arxiv.org/pdf/2511.xxxxx}
\end{links}

\section{Introduction} 
Image hashing has become a cornerstone of large-scale image retrieval systems due to its remarkable computational efficiency and compact storage footprint. The central objective is to map high-dimensional image data into low-dimensional binary codes that preserve semantic similarity in Hamming space. With the advent of deep learning, deep hashing has emerged as the dominant paradigm, significantly surpassing traditional shallow methods by learning powerful deep neural hash functions. This work focuses on the deep supervised hashing, where label information explicitly guides the learning of discriminative binary codes.
 
Existing deep supervised hashing methods can be broadly categorized by their similarity modeling strategies into three groups: \textbf{pairwise}, \textbf{triplet}, and \textbf{pointwise} approaches. Pairwise~\citep{Wu-dpsh-ijcai-2016, Zhangjie-hashnet-iccv-2017} and triplet~\citep{Hanjiang-dnnh-cvpr-2015, Xiaofang-dtsh-accv-2016} methods aim to preserve local similarity relationships among pairs or triplets of samples, often incurring quadratic or higher computational complexity with respect to the number of samples. In contrast, pointwise methods~\citep{Huei-ssdh-pami-2018, Shupeng-greedyhash-nips-2018, Li-csq-cvpr-2020} directly leverage class labels, achieving linear complexity.

Recently, pointwise approaches based on \textbf{hash centers} have attracted increasing attention~\citep{Li-csq-cvpr-2020, Jiun-orthoHash-nips-2021, Liangdao-shc-accv-2022, Liangdao-mds-cvpr-2023, Xingming-VTDH-KBS-2024}. These methods predefine a binary hash center for each class and train networks to align image representations with their assigned centers. While state-of-the-art models such as CSQ~\citep{Li-csq-cvpr-2020}, OrthoHash~\citep{Jiun-orthoHash-nips-2021}, and MDS~\citep{Liangdao-mds-cvpr-2023} have achieved impressive results, they typically fix the assignment of hash centers to classes randomly at initialization. This overlooks inter-class semantic correlations—e.g., semantically similar classes like ``cat'' and ``dog'' should ideally have closer hash centers than unrelated classes like ``cat'' and ``car''—limiting their ability to capture global semantic structures.

To mitigate this, SHC~\citep{Li-SHC-TOIS-2025} adopts a two-stage strategy, first generating semantically aware hash centers via classifier-based similarity estimation and iterative optimization, then learning the hash function. Although effective, this approach introduces heavy computational overhead, breaks end-to-end trainability, and relies on similarity measures rooted in classification objectives, which may not align perfectly with retrieval goals.

To address these issues, we propose \textbf{Center-Reassigned Hashing (CRH)}, a novel framework that dynamically optimizes hash center assignments during hash function training. Unlike existing methods that fix center assignments after initialization, CRH iteratively reassigns centers throughout training, progressively aligning them with inter-class semantic relationships. Concretely, CRH comprises three key components: (1) hash codebook initialization, followed by alternating stages of (2) hash function optimization and (3) hash center reassignment. This design eliminates the need for explicit pre-training or offline hash center generation, enabling direct end-to-end learning. Crucially, our ``reassignment'' refers to refining class-to-center assignments, not altering the binary codebook itself.

To sum up, our contributions are as follows: 
\begin{itemize}
    \item We introduce a novel \textit{reassignment-based mechanism} for updating hash centers, enabling the joint learning of hash function and semantic hash centers without requiring separate pre-training or extra hash center generation phase.
    
    \item We propose a flexible \textit{hash codebook} consisting of $M (\ge C)$ candidate hash centers, along with \textit{a multi-head design} (e.g., $H$-head) that effectively expands the codebook capacity from $M$ to $M^H$, allowing finer-grained semantic representations without increasing the physical codebook size $M$.

    \item Rich experiments on single-label and multi-label benchmarks demonstrate that our CRH method can effectively learns semantically meaningful hash centers and achieve consistent superiority over existing state-of-the-art deep hashing methods. 
\end{itemize}

\section{Related Work}

Hashing approaches can be broadly categorized into data-independent~\citep{Piotr-lsh-stoc-1998, Aristides-lsh2-vldb-1999, Moses-lsh3-stoc-2002} and data-dependent methods~\citep{Yair-sh-nips-2008, Brian-bre-nips-2009, Mohammad-mlh-icml-2011, Weihao-ih-nips-2012, Wei-ksh-cvpr-2012, Yunchao-itq-pami-2013}. The latter, particularly deep hashing methods~\citep{Wu-dpsh-ijcai-2016, Qi-DSDH-nips-2017, Qing-dcmh-cvpr-2017, Qing-adsh-aaai-2018, Jiun-orthoHash-nips-2021, Jianglin-dsah-isci-2022, Liangdao-shc-accv-2022, Liangdao-mds-cvpr-2023}, have become dominant due to their ability to learn compact and discriminative hash codes. Deep hashing techniques can be further classified into pairwise, triplet, and pointwise methods based on their supervision paradigms.

\textbf{Pairwise and Triplet Methods} learn hash functions by preserving local similarity structures. Pairwise approaches \citep{Wu-dpsh-ijcai-2016, Zhangjie-hashnet-iccv-2017, Yue-dch-cvpr-2018} minimize/maximize distances between similar/dissimilar pairs, while triplet methods \citep{Hanjiang-dnnh-cvpr-2015, Xiaofang-dtsh-accv-2016} enforce relative distance constraints. Although effective, these methods suffer from $\mathcal{O}(N^2)$ ($N$ is the number of data samples)  or more computational complexity and struggle to capture global data structures~\citep{Li-csq-cvpr-2020}.

\textbf{Pointwise Methods} overcome these limitations by utilizing direct label supervision. Early approaches \citep{Huei-ssdh-pami-2018, Shupeng-greedyhash-nips-2018} treated hashing as a classification problem. More recent center-based methods \citep{Li-csq-cvpr-2020, Lixin-dpn-ijcai-2020, Jiun-orthoHash-nips-2021, Liangdao-mds-cvpr-2023} assign predefined hash centers to each class, achieving $\mathcal{O}(NC)$ ($C$ denotes the number of classes) complexity and better global similarity preservation. However, these approaches typically generate centers through random sampling (CSQ) or combinatorial optimizations (MDS), ignoring inter-class semantic relationships.
 
Recent work \citep{Liangdao-shc-accv-2022,Li-SHC-TOIS-2025} proposed a two-stage method that first injects semantics into hash centers via class relationship estimation and discrete optimization, then learns hash function with these fixed centers.
While effective, these approaches require solving NP-hard optimization problems and suffer from stage-wise optimization gaps. In contrast, our CRH framework dynamically adjusts hash centers during end-to-end training, naturally encoding semantic relationships. 

\section{The Proposed Method}
\begin{figure}[ht]
    \centering
    \includegraphics[width=1.00\columnwidth]{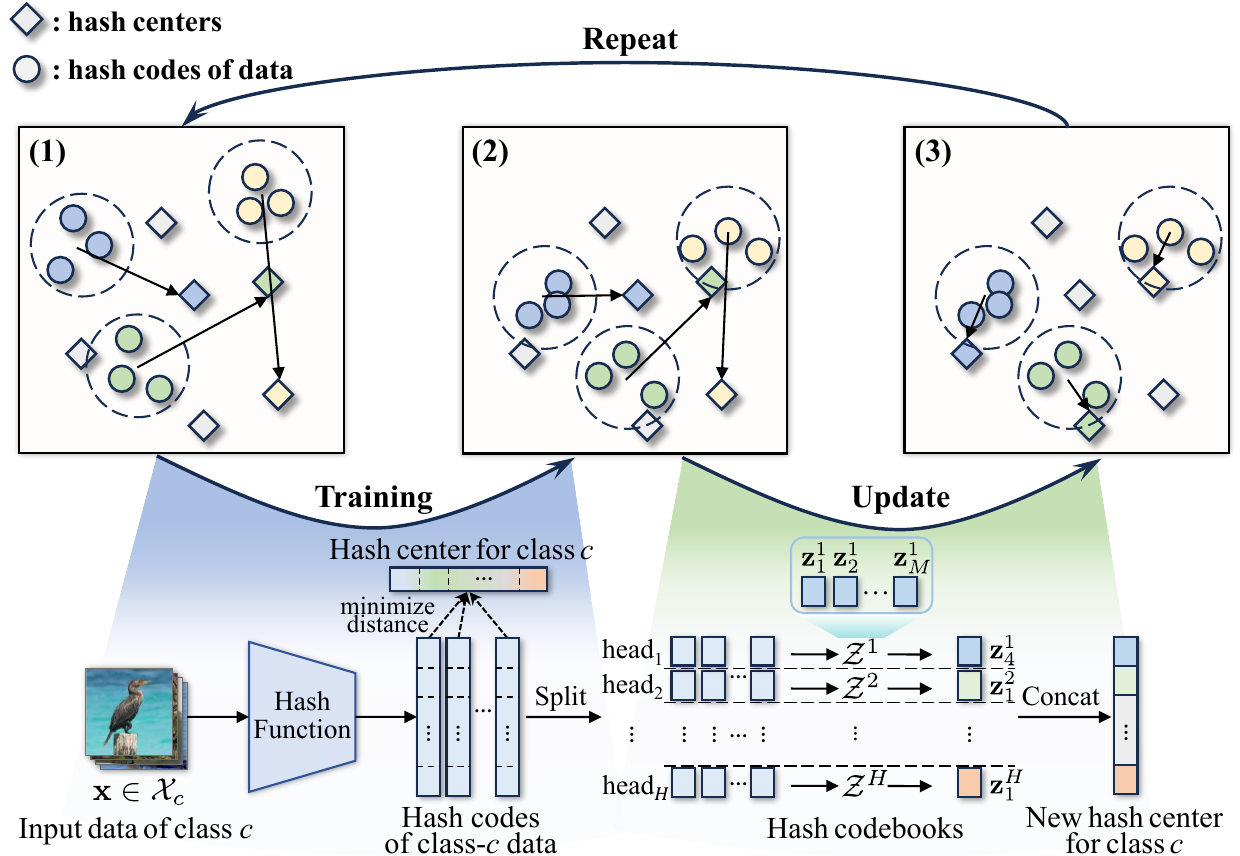} 
    \caption{
The overall framework of CRH.
\textbf{Top}: Hamming space visualization of the iterative hash center reassignment across 3 stages: (1) initial or previous assignment, (2) hash code convergence, and (3) updated center assignment. Three colors represent three classes.
\textbf{Bottom}: hash function training and multi-head update process for class $c$, where each head independently updates its sub-center $\mathbf{z}_m^h$ on a hash split, followed by concatenation into the full center $\mathbf{c}_c$.
}
    \label{fig_fram}
\end{figure}

\subsection{Problem Statement}
Given a dataset \( \mathcal{X} = \{(\mathbf{x}_n, \mathbf{y}_n)\}_{n=1}^N \) with \( N \) samples, each feature vector \( \mathbf{x}_n \in \mathbb{R}^D \) is associated with a multi-hot label vector \( \mathbf{y}_n=[y_{n1}, \cdots, y_{nC}]^\top \in \{0, 1\}^C \). Here, \( C \) denotes the total number of classes, and \( y_{nc} = 1 \) indicates that the  \( n \)-th sample belongs to the \( c \)-th class, while \( y_{nc} = 0 \) otherwise. The objective of supervised hashing is to learn an encoding function \( \text{f}: \mathbb{R}^D \to \{-1, 1\}^K \) that maps an input \( \mathbf{x} \) to a compact \( K \)-bit binary hash code \( \text{f}(\mathbf{x}) \). This function should preserve semantic similarity, ensuring that the Hamming distance between the hash codes of samples sharing at least one common category is smaller than that of samples with no overlapping categories.  

\subsection{Overall Framework}

Figure~\ref{fig_fram} illustrates the overall framework of CRH.
(1) We begin by randomly constructing a codebook $\mathcal{Z}$ comprising $M$ binary vectors of length $K$, where $M$ exceeds the number of classes $C$. We then randomly select $C$ vectors from $\mathcal{Z}$ to serve as the initial hash centers for the $C$ classes.
(2) Given these class-specific hash centers, we train a hash function that maps input samples to their binary hash codes.
(3) After training, we update the class hash centers by reassigning them using all training samples' current hash codes and the codebook:
specifically, we identify $C$ vectors from the codebook $\mathcal{Z}$ that minimize the total distance between the training samples’ hash codes and their nearest class centers, effectively yielding a new set of hash centers.
(4) Steps (2) and (3) are iteratively repeated until convergence, enabling the simultaneous learning of semantically structured hash centers and the hash function.
(5) Notably, inspired by multi-head attention, during the center update process we partition the codebook $\mathcal{Z}$ into $H$ sub-codebooks $\{\mathcal{Z}^h\}_{h=1}^H$. Each sub-codebook independently produces a sub-hash center $\mathbf{z}_m^h$ ($m\in \{1,\cdots,M\}$, $h \in \{1,\cdots,H\}$), and the final hash center $\mathbf{c}_c$ ($c \in \{1,\cdots,C\}$) for each class is formed by concatenating these sub-hash centers.

The detailed procedures for steps (1), (2), and (3) are described as follows.

\subsection{Initialization of Hash Centers}

In contrast to existing hash-center-based methods that assign one hash center per category (i.e., resulting in exactly 
$C$ hash centers for a $C$-class dataset), we propose to construct an expanded pool of candidate hash centers. Specifically, we define a \textit{hash codebook} \(\mathcal{Z} = \{\mathbf{z}_m\}_{m=1}^M\) consisting of $M$ binary hash codes \(\mathbf{z}_m \in \{-1, 1\}^K\) ($M \ge C$). This codebook serves as a candidate set from which class-specific hash centers are selected—each class center must be assigned a distinct element from $\mathcal{Z}$.

$\mathcal{Z}$ is initially populated by sampling from the Bernoulli distribution Bern(0.5)~\citep{Li-csq-cvpr-2020}. In this approach, each bit of every hash code is independently set to $-1$ or $1$ with equal probability. However, when $K$ is small (e.g., $K\leq16$) and $M$ is large, this independent sampling may lead to duplicate hash codes within $\mathcal{Z}$, reducing the effective diversity of the codebook.

To address this problem, we instead uniformly sample $M$ unique binary vectors from the full space $\{-1,1\}^K\!$ to construct $\mathcal{Z}$ for small $K$, of which the procedure is computationally efficient.
Moreover, under this uniform sampling strategy, the expected average Hamming distance between any two hash codes in $\mathcal{Z}$ is at least $K/2$~\citep{Li-csq-cvpr-2020}, ensuring sufficient separation among the candidate centers.

Once the hash codebook $\mathcal{Z}$ is constructed, we randomly select $C$ distinct elements from it to serve as the initial class-specific hash centers $\{\mathbf{c}_c\}_{c=1}^C$, where $\mathbf{c}_c$ denotes the hash center assigned to the $c$-th class. These assignments will be dynamically refined during training through our proposed center reassignment mechanism.

\subsection{Training of Hash Function}
Given the initialized or updated hash centers, we train the hash function $\text{f}(\mathbf{x})$, implemented as a deep neural network (e.g., a pre-trained ResNet-34 backbone followed by a linear projection layer with a $\tanh$ activation), to map each input sample $\mathbf{x}_n$ to a continuous binary-like code $\mathbf{h}_n=\tanh(\mathbf{v}_n) \in (-1, 1)^K$, where $\mathbf{v}_n$ denotes the output of the linear layer. For retrieval, these codes are subsequently binarized as $\mathbf{b}_n=\text{sign}(\mathbf{h}_n)$.

We adopt a margin-based cross-entropy loss \( \mathcal{L}_\mathrm{CE} \)~\citep{Jiun-orthoHash-nips-2021}, defined as:
{\small
\begin{equation}
    \mathcal{L}_\mathrm{CE} = -\frac{1}{N} \sum_{n=1}^N \sum_{c=1}^C \frac{y_{nc}}{\|\mathbf{y}_n\|_1} \log p_{nc},
\end{equation}
}
with softmax probability:
{\small
\begin{equation}
    p_{nc} = \frac{\exp(s \cdot \text{sim}(\mathbf{v}_n, \mathbf{c}_c))}{\sum_{j=1}^C \exp(s \cdot \text{sim}(\mathbf{v}_n, \mathbf{c}_j))},
\end{equation}
}
where \(
\text{sim}(\mathbf{v}_n, \mathbf{c}_c) = \mathbf{v}_n^\top \mathbf{c}_c/(\|\mathbf{v}_n\|_2 \cdot \|\mathbf{c}_c\|_2) - y_{nc} \cdot margin
\) incorporates cosine similarity with a margin \( margin \), \( s \) is a scaling factor, and {\small\( \|\mathbf{y}_n\|_1 = \sum_{c=1}^C |y_{nc}| \)} counts the number of labels for the \( n \)-th data point. This loss pulls embeddings toward their corresponding hash centers while pushing them away from others. 

To minimize quantization errors due to relaxations in the \(\tanh\) layer, we introduce a quantization loss:
{\small
\begin{equation}
    \mathcal{L}_q = 
    \frac{1}{NK} \sum_{n=1}^{N} \left\| \text{abs}(\mathbf{h}_n) - \mathbf{1} \right\|_2^2 =
    \frac{1}{NK} \sum_{n=1}^{N}\sum_{k=1}^{K}(|h_{nk}|-1)^2,
\end{equation} 
}
where \( \text{abs}(\cdot) \) denotes the element-wise absolute value, \( \mathbf{1} \) is a vector of all ones, and {\small\( \|\cdot\|_2 \)} is the \( \ell_2 \)-norm.  

The overall objective is:  
{\small
\begin{equation}\label{eq:loss}
    \mathcal{L} = \mathcal{L}_\mathrm{CE} + \lambda \mathcal{L}_q,
\end{equation}
}
where \( \lambda (\geq 0) \) balances $\mathcal{L}_\mathrm{CE}$ and $\mathcal{L}_q$.

\subsection{Updating of Hash Centers via Reassignment}
The initial assignment of hash centers to classes is random and therefore semantically agnostic. 
To enhance semantic alignment, we dynamically reassign hash centers throughout training by leveraging evolving data representations without directly numerically optimizing the centers themselves.
The proposed reassignment procedure is detailed below.

\textbf{Reassignment Strategy.}
At the end of selected training epochs (e.g., for every five epochs), we update the hash centers by first passing the training data through the hash function $\text{f}(\mathbf{x})$ to produce continuous representations $\mathbf{h}_\mathbf{x}$, which are subsequently binarized to $\text{sign}(\mathbf{h}_\mathbf{x})$. For each class $c$, let \( \mathcal{X}_c \) denote its collection of data samples. We compute the mean Euclidean distance between $\text{sign}(\mathbf{h}_\mathbf{x})$ and each candidate center $\mathbf{z}_m \in \mathcal{Z}$ as the error of assigning $\mathbf{z}_m$ to class $c$, yielding a cost matrix $\mathbf{L} = (l_{cm})_{C \times M}$ defined by:
{\small
\begin{equation}\label{eq:lij-single-label}
    l_{cm} = \frac{1}{|\mathcal{X}_c|} \sum_{\mathbf{x} \in \mathcal{X}_c} \left\| \text{sign}(\mathbf{h}_\mathbf{x}) - \mathbf{z}_m \right\|^2_2,
\end{equation}
}
where \( c = 1, 2, \cdots, C \), \( m = 1, 2, \cdots, M \), and \( |\mathcal{X}_c| \) denotes the number of data samples in category \( c \). 

We then seek an optimal one-to-one mapping from classes to codebook elements that minimizes the total assignment error:
{\small
\begin{equation}
    (j_1^*, j_2^*, \cdots, j_C^*) = \mathop{\arg\min}_{\substack{j_1, \cdots, j_C \in \{1, \cdots, M\} \\ j_i \neq j_k \text{ for } i \neq k}} \sum_{c=1}^C l_{c j_c},  \end{equation}}
where \( j_c^* \) represents the index of the center in \( \mathcal{Z} \) assigned to the \( c \)-th category in the optimal solution. This combinatorial optimization problem can be efficiently solved using the Hungarian algorithm~\citep{kuhn-hungarian-nrlq-1955}. 

While the Hungarian algorithm provides an optimal assignment within each epoch, empirical results reveal that a \textit{greedy algorithm} performs better overall. It sequentially assigns each class to its best unassigned center in $\mathcal{Z}$, following a random class order. This stochasticity may lead to suboptimal per-epoch assignments but helps prevent overfitting to transient local minima that shift during training.

\textbf{Multi-Label Handling.}
In multi-label settings, the supervision for each sample reflects a mixture of categories, diluting the representativeness for any single class.
To address this, we follow DCSH~\citep{Abin-dcsh-icassp-2022} by weighting each data point's contribution with factor \( 1 / \|\mathbf{y}_\mathbf{x}\|_1 \), where \( \|\mathbf{y}_\mathbf{x}\|_1 \) denotes the number of labels associated with sample \( \mathbf{x} \). The weighted error for class \(c\) is:
{\small
\begin{equation}\label{eq:lij}
    l_{cm} = \frac{1}{\sum_{\mathbf{x} \in \mathcal{X}_c} \frac{1}{\|\mathbf{y}_\mathbf{x}\|_1}} \sum_{\mathbf{x} \in \mathcal{X}_c} \frac{1}{\|\mathbf{y}_\mathbf{x}\|_1} \left\| \text{sign}(\mathbf{h}_\mathbf{x}) - \mathbf{z}_m \right\|^2_2.
\end{equation}
}
By incorporating these weights, data points with more category labels contribute less to the computation of \( l_{cm} \), ensuring that the cost matrix \( \mathbf{L} \) more accurately reflects the distance relationships between the data of different categories and the elements in \( \mathcal{Z} \). For single-label datasets, Eq.~(\ref{eq:lij}) naturally reduces to Eq.~(\ref{eq:lij-single-label}).

\textbf{Multi-Head Codebook Design.}
To amplify the representational power of the hash codebook $\mathcal{Z}$ without increasing its cardinality $M\!$, we split each $K\!$-dimensional vector $\mathbf{z}_m\!$ into $H\!$ heads (i.e., $\{\mathbf{z}^1_m,\!\cdots\!,\mathbf{z}^H_m\}$), with each $\mathbf{z}^h_m\!$ sized by $\!{d\!=\!K/H}$. Each head independently performs center reassignment using its sub-codebook $\!{\mathcal{Z}^h\!=\!\{\mathbf{z}_1^h,\!\cdots\!,\mathbf{z}_M^h\}}$, derived from the original codebook $\!\mathcal{Z}$. Specifically, for head $h$, the corresponding components $\mathbf{h}^h_\mathbf{x}$ and sub-codebook $\mathcal{Z}^h$ are used to compute a per-head cost matrix $\mathbf{L}^h$, followed by greedy assignment to yield updated centers $\{\mathbf{c}^h_c\}_{c=1}^C$. Final centers are obtained by concatenating the head-specific centers:
{\small
\begin{equation}
    \mathbf{c}_c = \text{concat}(\mathbf{c}_c^1, \cdots, \mathbf{c}_c^H).
\end{equation}
}

This multi-head design effectively enlarges the codebook capacity from $M$ to $M^H$, enabling richer semantic discrimination. Since $d$ bits can represent at most $2^d$ distinct binary vectors, we constrain  $H \leq K / \log_2 M$  to ensure the $M$ elements within each $\mathcal{Z}^h$ remain mutually distinct and avoid collisions. Algorithm~\ref{alg:algorithm_full} summarizes the overall procedure.
 
\subsection{Optimization Perspective of CRH} 
From an optimization standpoint, center-based hashing inherently poses an NP-hard problem: determining both a $C$-element subset from $\{-1, 1\}^K$ and its assignment to categories that most effectively encapsulates inter-class semantic relationships, thereby maximizing retrieval performance. Our CRH method offers an efficient approximation by introducing two pivotal mechanisms: (1) \textbf{Codebook Constraint}: Instead of searching the entire Hamming space, we restrict the selection of hash centers to a predefined codebook \( \mathcal{Z} \subset \{-1, 1\}^K \) with cardinality $|\mathcal{Z}| \ll 2^K$, significantly reducing the solution space. (2) \textbf{Dynamic Update}: We iteratively refine the hash centers in tandem with hash function training, thereby circumventing the combinatorial explosion associated with exhaustively exploring all possible configurations.

\begin{algorithm}[t]
\caption{Center-Reassigned Hashing (CRH)}
\label{alg:algorithm_full}
\textbf{Input:} $K$, $C$, \& ($M$, $\lambda$, $H$, $s$, $margin$). \\
\textbf{Output:} $\text{f}(\cdot)$ \& $\{\mathbf{c}_c\}_{c=1}^C$.
\begin{algorithmic}[1]
\STATE Generate a hash codebook $\mathcal{Z}$ by randomly sampling $M$ binary vectors of dimension $K$;
\STATE Sample $C$ binary vectors from $\mathcal{Z}$ as initial class centers;

\REPEAT
\STATE Train the hash function $\text{f}(\cdot)$ by minimizing the objective in Eq.~(\ref{eq:loss}) for a specific number of epochs;
\FOR{$h=1$ to $H$}
\STATE Compute the cost matrix $\mathbf{L}^h$ using Eq.~(\ref{eq:lij});
\STATE Reassign sub-centers $\mathbf{c}_c^h$ via the greedy algorithm;
\ENDFOR
\UNTIL{\textit{convergence or max epochs reached};}

\STATE \textbf{Return} the hash function $\text{f}(\cdot)$ and centers $\{\mathbf{c}_c\}_{c=1}^C$.
\end{algorithmic}
\end{algorithm}

\section{Experiment}

\subsection{Experimental Setups}

\subsubsection{Datasets}
We evaluate CRH on three widely used datasets covering both single- and multi-label retrieval tasks:

    \textbf{Stanford Cars}~\cite{Jonathan-cars196-iccvw-2013}: a single-label dataset with 196 vehicle categories, containing 8,144 training and 8,041 test images. Following~\citep{Liangdao-mds-cvpr-2023}, we use the training set for both model learning and the retrieval database, and the test set as queries.

    \textbf{NABirds}~\citep{Grant-nabirds-cvpr-2015}: a fine-grained single-label dataset with 555 bird species, split into 23,929 training and 24,633 test images. We adopt the same training/retrieval and query split as for Stanford Cars.

    \textbf{MS COCO}~\citep{Tsung-mscoco-eccv-2014}: a multi-label dataset with 80 object categories. Following~\citep{Zhangjie-hashnet-iccv-2017}, we sample 5,000 query images, use 10,000 for training, and assign the remaining images to the retrieval database.

\begin{table*}[ht!]\scriptsize
    \centering
    \setlength{\tabcolsep}{11.5pt}
    \begin{tabular}{l|ccc|ccc|ccc}
        \Xhline{0.8pt}
        \multirow{2}{*}{Methods} & \multicolumn{3}{c|}{\textbf{Stanford Cars} (mAP@all)} & \multicolumn{3}{c|}{\textbf{NABirds} (mAP@all)} & \multicolumn{3}{c}{\textbf{MS COCO} (mAP@5K)} \\
        \cline{2-10}
        & 16 bits & 32 bits & 64 bits & 16 bits & 32 bits & 64 bits & 16 bits & 32 bits & 64 bits \\
        \hline       
        DTSH~\citep{Xiaofang-dtsh-accv-2016} & 57.48 & 71.32 & 76.20 & 23.46 & 32.91 & 44.01 & \underline{79.14} & 81.66 & 82.79 \\
        HashNet~\citep{Zhangjie-hashnet-iccv-2017} & 25.99 & 40.73 & 50.84 & 11.84 & 20.26 & 29.47 & 70.62 & 74.36 & 76.92 \\
        GreedyHash~\citep{Shupeng-greedyhash-nips-2018} & 80.95 & 86.27 & 87.05 & 64.10 & 71.14 & 72.86 & 75.05 & 81.13 & 84.52 \\
        CSQ~\citep{Li-csq-cvpr-2020} & 79.13 & 85.77 & 87.17 & 65.70 & 70.62 & 74.07 & 72.44 & 81.32 & 85.30 \\
        OrthoHash~\citep{Jiun-orthoHash-nips-2021} & 83.97 & 86.38 & 87.35 & \underline{69.86} & \underline{71.95} & 73.54 & 78.98 & \underline{85.01} & \underline{87.07} \\
        MDS~\citep{Liangdao-mds-cvpr-2023} & \underline{85.33} & 86.49 & 87.76 & 65.14 & 71.47 & 74.08 & 73.53 & 79.98 & 82.57 \\
        SHC~\cite{Li-SHC-TOIS-2025} & 83.21 & \underline{86.93} & \underline{88.43} & 64.58 & 71.89 & \underline{74.15} & 73.43 & 79.33 & 83.85 \\
        \hline
        CRH (Ours) & \textbf{87.31} & \textbf{89.20} & \textbf{90.27} & \textbf{74.01} & \textbf{76.69} & \textbf{77.71} & \textbf{82.71} & \textbf{85.79} & \textbf{87.44} \\
        \Xhline{0.8pt}
    \end{tabular}
    \caption{Comparison of retrieval performance (mAP, \%) between our method and deep hashing baselines on three datasets at multiple bit lengths. Note that the mAP values in \textbf{Bold} and \underline{underlined} indicate the best and second-best results, respectively.}
    \label{tab:mAP}
\end{table*}

\begin{table*}[ht!]\scriptsize
    \centering
    \setlength{\tabcolsep}{15.5pt}
    \begin{tabular}{l|ccc|ccc|ccc}
        \Xhline{0.8pt}
        \multirow{2}{*}{Methods} & \multicolumn{3}{c|}{\textbf{Stanford Cars} (mAP@all)} & \multicolumn{3}{c|}{\textbf{NABirds} (mAP@all)} & \multicolumn{3}{c}{\textbf{MS COCO} (mAP@5K)} \\
        \cline{2-10}
        & 16 bits & 32 bits & 64 bits & 16 bits & 32 bits & 64 bits & 16 bits & 32 bits & 64 bits \\
        \hline       
        CRH   & \textbf{87.31} & \textbf{89.20} & \textbf{90.27} & \textbf{74.01} & \textbf{76.69} & \textbf{77.71} & \textbf{82.71} & \textbf{85.79} & \textbf{87.44} \\
        CRH-M & \textbf{87.31} & \underline{88.31}   & \underline{88.73}    & \textbf{74.01} & \underline{74.98}    & \underline{75.94}    & \underline{81.60}     & \underline{85.45}    & 86.75          \\
        CRH-U & \underline{85.12}          & 87.22         & 87.10           & \underline{70.69}          & 73.06          & 74.51          & 78.45          & 84.20           & \underline{86.95}   \\
        \Xhline{0.8pt}
    \end{tabular}
    \caption{Comparison of retrieval performance (mAP, \%) between CRH and its ablated variants (CRH-M, CRH-U).}
    \label{tab:ACRH}
\end{table*}

\subsubsection{Metrics and Baselines}
We evaluate retrieval using mean Average Precision (mAP), reporting mAP@all for Stanford Cars and NABirds, and mAP@5,000 for MS COCO. Our method is compared against seven state-of-the-art deep hashing baselines: four center-based methods (CSQ~\citep{Li-csq-cvpr-2020}, OrthoHash~\citep{Jiun-orthoHash-nips-2021}, MDS~\citep{Liangdao-mds-cvpr-2023}, SHC~\cite{Li-SHC-TOIS-2025}) and three general approaches (HashNet~\citep{Zhangjie-hashnet-iccv-2017}, DTSH~\citep{Xiaofang-dtsh-accv-2016}, GreedyHash~\citep{Shupeng-greedyhash-nips-2018}). For fairness, all use the same pre-trained ResNet-34~\citep{Kaiming-resnet-cvpr-2016} backbone.

\subsubsection{Implementation Details}
We optimize the model using Adam~\citep{Diederik-adam-iclr-2015} with \((\beta_1, \beta_2)\!=\!(0.5, 0.999) \) and a weight decay of \( 10^{-5}\!\). The margin \(\!margin\!\) is set to $0.2\!$ \citep{Jiun-orthoHash-nips-2021}, and the scale factor to \(s\!=\!\sqrt{2}\log{(C\!-\!1)}\!\)~\citep{Xiao-adacos-cvpr-2019}. The codebook size is fixed at \({M\!=\!2C\!}\) (as shown in Figure~\ref{fig_hyparam}). We employ a cosine annealing schedule~\citep{Ilya-sgdr-iclr-2017} with an initial learning rate of \( 10^{-4}\!\). The hyperparameter \( \lambda \) is set to \( 0.1 \) for Stanford Cars and NABirds, and $0$ for MS COCO (with more details in the \textit{Supplementary}). The head dimension \( d \) is set to $16$, $16$, and $8$ for the three datasets, respectively (as shown in Figure~\ref{fig_hyparam}). By default, hash center reassignment uses the greedy algorithm, applied every epoch for the first 20 epochs and every 5 epochs thereafter (as shown in Figure~\ref{fig_ui}). Training runs for 300 epochs on the single-label datasets and 30 epochs on MS COCO, with a batch size of 128. 

\subsection{Results}
The mAP results are summed in Table~\ref{tab:mAP}. Our CRH method consistently achieves state-of-the-art performance across all datasets, surpassing the strongest baselines by relative margins of 2.1\%–2.6\% (Stanford Cars), 4.8\%–6.6\% (NABirds), and 0.4\%–4.5\% (MS COCO). The substantial improvement on NABirds—a dataset characterized by intricate inter-class relationships due to its large category count—demonstrates the efficacy of our hash center update mechanism in modeling fine-grained semantic structures. On Stanford Cars, CRH delivers robust performance gains across all hash code lengths. For MS COCO, the most significant improvements occur at 16 bits, with diminishing returns observed for longer codes. Notably, DTSH and HashNet exhibit markedly inferior performance on single-label datasets with hundreds of categories, suggesting their tuple-similarity-based optimization objectives are less effective at capturing inter-class relationships in large-scale scenarios.

\subsection{Ablation Studies}
We conduct ablation studies to assess the contributions of two main components: center reassignment and the multi-head architecture. Specifically, we evaluate:
\begin{itemize}
    \item \textbf{CRH-U}: Disables center reassignment and consequently the multi-head mechanism, resulting in a fixed-center approach similar to CSQ/OrthoHash; 

    \item \textbf{CRH-M}: Retains reassignment but removes the multi-head mechanism (i.e., uses a single head, $H = 1$).
\end{itemize}
As illustrated in Table~\ref{tab:ACRH}, CRH-M consistently outperforms CRH-U across nearly all settings, achieving average relative mAP gains of 1.9\%, 3.08\%, and 1.76\% on Stanford Cars, NABirds, and MS COCO, respectively. These results underscore the effectiveness of dynamic center updates in capturing data semantics. Building on this, CRH achieves further improvements over CRH-M, with additional gains of 0.91\%, 1.54\%, and 0.85\% on the same datasets, demonstrating the benefit of the multi-head design in refining hash centers for enhanced representation quality.

To test generalizability, we incorporate our update mechanism into CSQ, MDS, and OrthoHash (denoted $(\cdot)_U$), keeping their original codebooks (size $C$) without multi-heads. Table~\ref{tab:Absl} shows consistent improvements in most cases, underscoring the broad applicability of our strategy for enhancing center-based hashing.
 
\begin{table*}[ht!]\scriptsize
    \centering
    \setlength{\tabcolsep}{14.6pt}
        \begin{tabular}{l|ccc|ccc|ccc}
            \Xhline{0.8pt}
            \multirow{2}{*}{Methods} & \multicolumn{3}{c|}{\textbf{Stanford Cars} (mAP@all)} & \multicolumn{3}{c|}{\textbf{NABirds} (mAP@all)} & \multicolumn{3}{c}{\textbf{MS COCO} (mAP@5K)} \\
            \cline{2-10}
            & 16 bits & 32 bits & 64 bits & 16 bits & 32 bits & 64 bits & 16 bits & 32 bits & 64 bits \\
            \hline       
            CSQ         & 79.13          & 85.77          & 87.17          & 65.70           & 70.62          & 74.07          & 72.44          & 81.32          & 85.30           \\
            CSQ$_U$       & \textbf{85.25} & \textbf{87.72} & \textbf{88.23} & \textbf{70.14} & \textbf{73.42} & \textbf{75.38} & \textbf{77.65} & \textbf{82.43} & \textbf{85.56}\\
            \hline 
            MDS         & 85.33          & 86.49          & 87.76          & 65.14          & 71.47          & 74.08          & 73.53          & 79.98          & 82.57          \\
            MDS$_U$       & \textbf{87.23} & \textbf{87.76} & \textbf{88.06} & \textbf{70.84} & \textbf{73.77} & \textbf{75.27} & \textbf{79.31} & \textbf{82.40}  & \textbf{83.13} \\
            \hline 
            OrthoHash   & 84.59          & 86.66          & \textbf{87.35}          & 69.80           & 72.13          & 73.54          & 78.98          & \textbf{85.01} & \textbf{87.07} \\
            OrthoHash$_U$ & \textbf{86.69} & \textbf{88.11} & \textbf{87.35} & \textbf{73.70}  & \textbf{74.30}  & \textbf{74.65} & \textbf{80.56} & 84.24          & 86.33  \\       
            \Xhline{0.8pt}
        \end{tabular}
    \caption{Performance comparison (mAP, \%) of hash-center-based methods with/without the proposed update mechanism.} 
    \label{tab:Absl}
\end{table*}

\begin{table*}[ht!]
    \centering
    \setlength{\tabcolsep}{7.2pt}
    \resizebox{\textwidth}{!}{
        \begin{tabular}{l|ccc|ccc|ccc}
            \Xhline{0.8pt}
            \multirow{2}{*}{Methods} & \multicolumn{3}{c|}{\textbf{Stanford Cars} (mAP@all)} & \multicolumn{3}{c|}{\textbf{NABirds} (mAP@all)} & \multicolumn{3}{c}{\textbf{MS COCO} (mAP@5K)} \\
            \cline{2-10}
            & 16 bits & 32 bits & 64 bits & 16 bits & 32 bits & 64 bits & 16 bits & 32 bits & 64 bits \\
            \hline       
            Seed   & 87.35 $\scriptstyle{\pm\,0.03}$ & 89.01 $\scriptstyle{\pm\,0.19}$ & 89.99 $\scriptstyle{\pm\,0.24}$ & 73.98 $\scriptstyle{\pm\,0.06}$ & 76.60 $\scriptstyle{\pm\,0.08}$ & 77.72 $\scriptstyle{\pm\,0.06}$ & 82.63 $\scriptstyle{\pm\,0.08}$ & 85.72 $\scriptstyle{\pm\,0.13}$ & 87.29 $\scriptstyle{\pm\,0.12}$ \\
            Init   & 87.46 $\scriptstyle{\pm\,0.11}$ & 88.89 $\scriptstyle{\pm\,0.27}$ & 90.23 $\scriptstyle{\pm\,0.06}$ & 74.09 $\scriptstyle{\pm\,0.20}$ & 76.62 $\scriptstyle{\pm\,0.17}$ & 77.83 $\scriptstyle{\pm\,0.21}$ & 82.75 $\scriptstyle{\pm\,0.03}$ & 85.74 $\scriptstyle{\pm\,0.09}$ & 87.19 $\scriptstyle{\pm\,0.20}$ \\
            Init-H & 87.23 $\scriptstyle{\pm\,0.10}$   & 88.80 $\scriptstyle{\pm\,0.18}$   & 89.45 $\scriptstyle{\pm\,0.06}$   & 73.45 $\scriptstyle{\pm\,0.19}$   & 76.03 $\scriptstyle{\pm\,0.02}$   & 77.53 $\scriptstyle{\pm\,0.02}$   & 82.23 $\scriptstyle{\pm\,0.31}$   & 85.73 $\scriptstyle{\pm\,0.07}$   & 87.08 $\scriptstyle{\pm\,0.10}$    \\     
            \Xhline{0.8pt}
        \end{tabular}
   }
    \caption{Performance impact of initialization randomness and update algorithms (mean±std mAP over 3 runs). ``Seed'': random update mechanism; ``Init'': varied center initializations; ``Init-H'': ``Init'' with Hungarian-algorithm updates.}
    \label{tab:random}
\end{table*}

\begin{table*}[ht!]
    \centering
    \resizebox{\textwidth}{!}{
        \begin{tabular}{l|ccc|ccc|ccc}
            \Xhline{0.8pt}
            \multirow{2}{*}{Methods} & \multicolumn{3}{c|}{\textbf{Stanford Cars}} & \multicolumn{3}{c|}{\textbf{NABirds}} & \multicolumn{3}{c}{\textbf{MS COCO}} \\
            \cline{2-10}
            & 16 bits & 32 bits & 64 bits & 16 bits & 32 bits & 64 bits & 16 bits & 32 bits & 64 bits \\
            \hline       
            Init   & 0.004       & -0.007      & -0.002       & -0.001       & 0.001        & 0.001        & 0.006       & -0.025      & -0.011      \\
            Learned   & 0.242       & 0.286       & 0.401        & 0.199        & 0.278        & 0.337        & 0.307       & 0.379       & 0.460        \\
            Learned-M & 0.242       & 0.185       & 0.115        & 0.199        & 0.138        & 0.102        & 0.157       & 0.101       & 0.135       \\
            Random    & 0.000 $\scriptstyle{\pm\,0.007}$ & 0.000 $\scriptstyle{\pm\,0.008}$ & -0.000 $\scriptstyle{\pm\,0.007}$ & -0.000 $\scriptstyle{\pm\,0.003}$ & -0.000 $\scriptstyle{\pm\,0.003}$ & -0.000 $\scriptstyle{\pm\,0.003}$ & 0.001 $\scriptstyle{\pm\,0.015}$ & 0.000 $\scriptstyle{\pm\,0.016}$ & 0.000 $\scriptstyle{\pm\,0.018}$    \\
            \hline 
            CSQ/OrthoHash & 0.005     & -0.005    & -0.008    & 0.004   & 0.000       & -0.001  & 0.002    & 0.008  & 0.000       \\
            MDS           & 0.002     & 0.004     & 0.006     & 0.001   & -0.002  & -0.002  & -0.002   & 0.010   & -0.002  \\
            SHC           & 0.022     & 0.046     & 0.011     & 0.001   & 0.011   & -0.004  & -0.002   & 0.010   & -0.002 \\ 
            \Xhline{0.8pt}
        \end{tabular}
    }
    \caption{Pearson correlation analysis between $\mathbf{S}^h$ and $\mathbf{S}^r$ under different center configurations: initialized (``Init''), learned (multi-head ``Learned''/single-head ``Learned-M''), and random assignments (``Random'', mean±std over 1,000 runs). Besides, the results of the four baseline methods (CSQ, OrthoHash, MDS, and SHC) are also included.}
    \label{tab:corr}
\end{table*}

\subsection{Robustness to Randomness}    
Our algorithm incorporates stochasticity through (1) random initialization of hash centers and (2) the greedy center update procedure. To assess their effects, we conduct:
\begin{itemize}
    \item \textbf{Initialization Robustness}: 3 runs with different random center initializations (``Init'').

    \item \textbf{Update Robustness}: 3 runs with different random seeds but identical initializations (``Seed'').
\end{itemize}
As shown in Table~\ref{tab:random}, the consistently low standard deviations confirm the algorithm’s stability against these sources of randomness.

We also compare our greedy update (``Init'') with the Hungarian algorithm (``Init-H''). The greedy algorithm provides clear advantages: (1) slightly higher mAP (relative gains of 0.41\%, 0.68\%, and 0.26\% across datasets), attributed to the stochasticity from its random class order which enables broader exploration of the solution space; and (2) lower computational complexity ($\mathcal{O}(HCM)$ vs. $\mathcal{O}(HC^2M)$).

\begin{figure*}[ht!]
    \centering
    \setlength{\tabcolsep}{0pt} 
    \begin{tabular}{@{}c@{\hspace{0.0475\textwidth}}c@{\hspace{0.0475\textwidth}}c@{}} 
        \raisebox{-\height}{\includegraphics[width=0.27\textwidth]{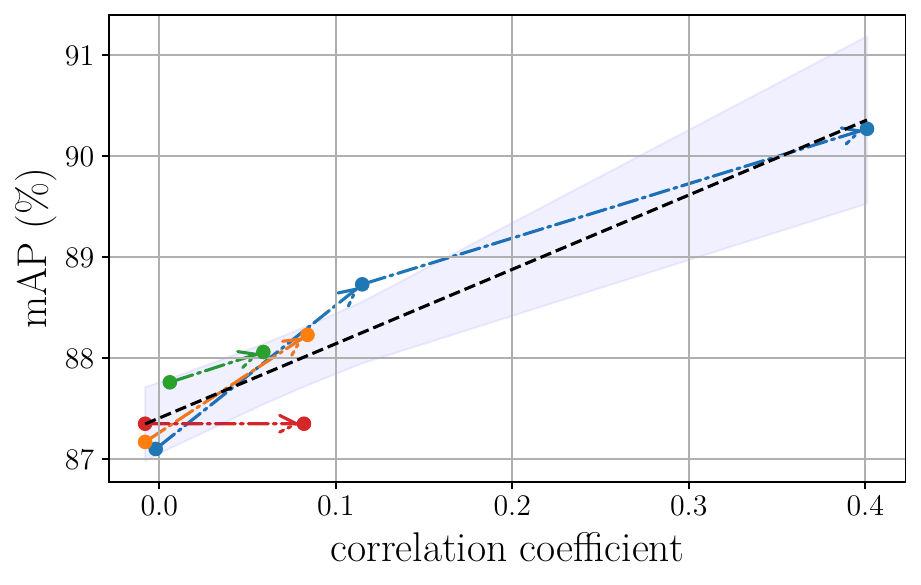}} &
        \raisebox{-\height}{\includegraphics[width=0.27\textwidth]{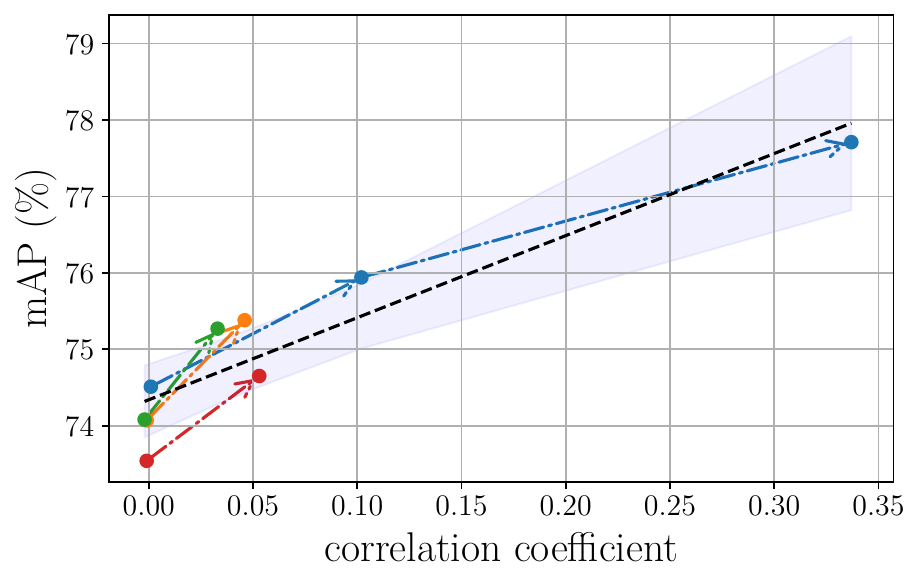}} &
        \raisebox{-\height}{\includegraphics[width=0.27\textwidth]{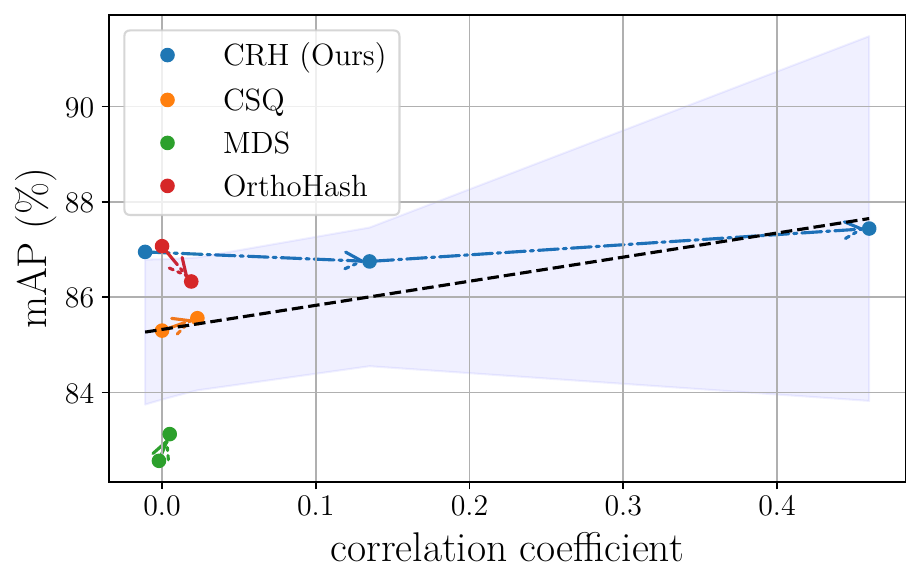}} \\
        
        \makebox[0.27\textwidth]{\centering (a) Stanford Cars} & 
        \makebox[0.27\textwidth]{\centering (b) NABirds} & 
        \makebox[0.27\textwidth]{\centering (c) MS COCO} \\
    \end{tabular}
    \caption{mAP vs. PCC (64-bit) on different datasets. Arrows link baseline variants (``original'' → ``updated'', e.g., MDS → MDS$_U$) and \textbf{CRH-U} → \textbf{CRH-M} → \textbf{CRH}. Regression lines with 95\% confidence intervals indicate the linear trend.}
    \label{fig_mvsc}
\end{figure*}

\begin{figure*}[t]
    \centering
    \begin{subfigure}{0.28\textwidth}
        \includegraphics[width=\linewidth]{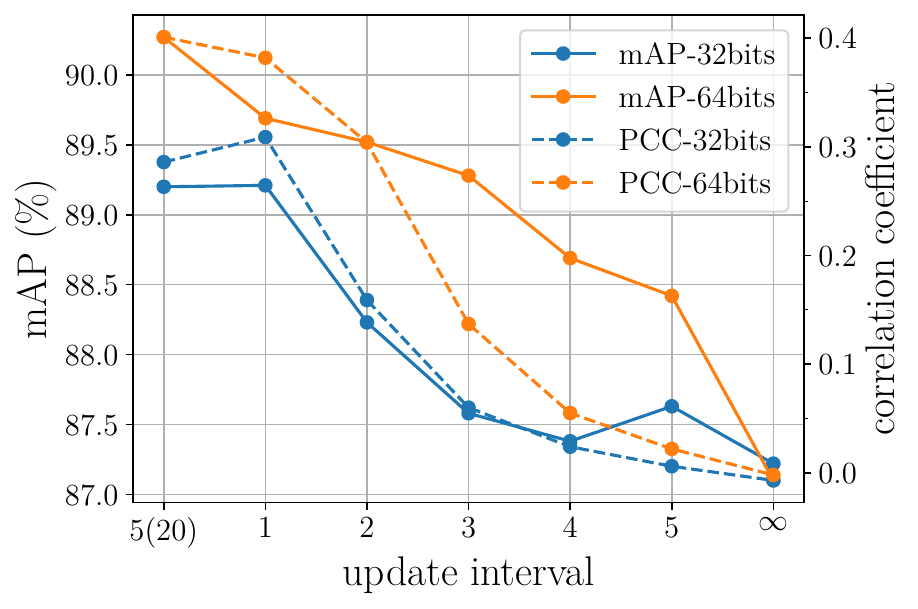}
        \caption{Stanford Cars}
        \label{subfig:cars}
    \end{subfigure}
    \hspace{0.04\textwidth}%
    \begin{subfigure}{0.28\textwidth}
        \includegraphics[width=\linewidth]{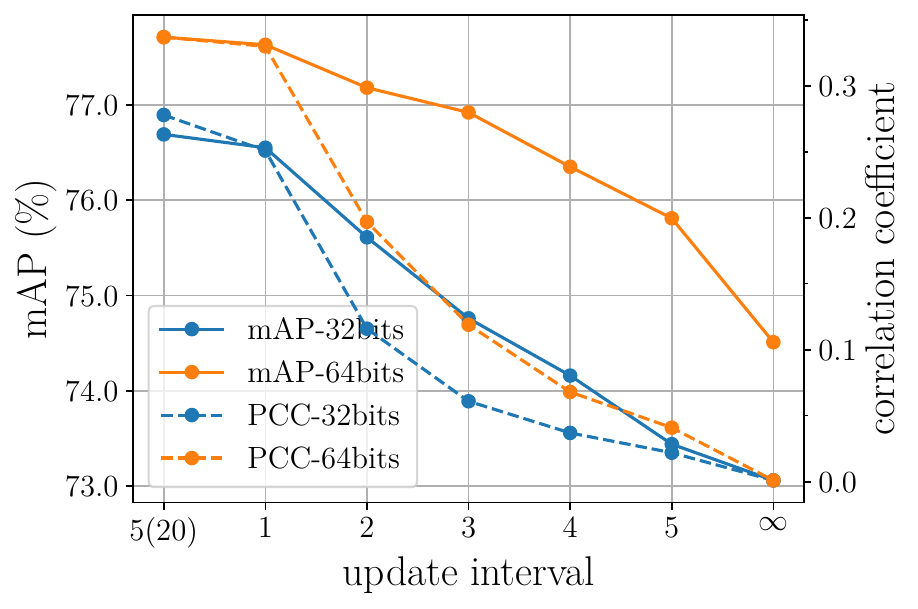}
        \caption{NABirds}
        \label{subfig:birds}
    \end{subfigure}
    \hspace{0.04\textwidth}%
    \begin{subfigure}{0.28\textwidth}
        \includegraphics[width=\linewidth]{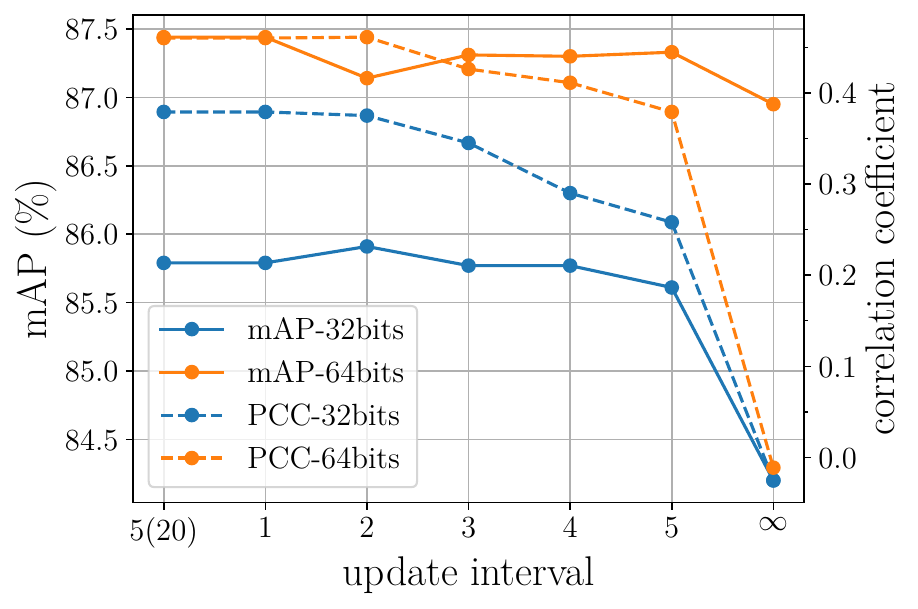}
        \caption{MS COCO}
        \label{subfig:coco}
    \end{subfigure}
    
    \caption{Impact of hash center update frequency on mAP/PCC across datasets. ``5 (20)'' indicates updates every epoch for the first 20 epochs, then every 5; ``interval=2'' denotes updates every 2 epochs (similarly for other values); ``$\infty$'' means no updates.}
    \label{fig_ui}
\end{figure*}

\begin{figure}[ht!]
    \centering
    \setlength{\tabcolsep}{0pt} 
    \begin{tabular}{@{} p{\dimexpr0.5\columnwidth-0.5pt} p{\dimexpr0.5\columnwidth-0.5pt} @{}}
        \begin{subfigure}[t]{\linewidth}
            \includegraphics[width=\linewidth, height=\textheight, keepaspectratio]{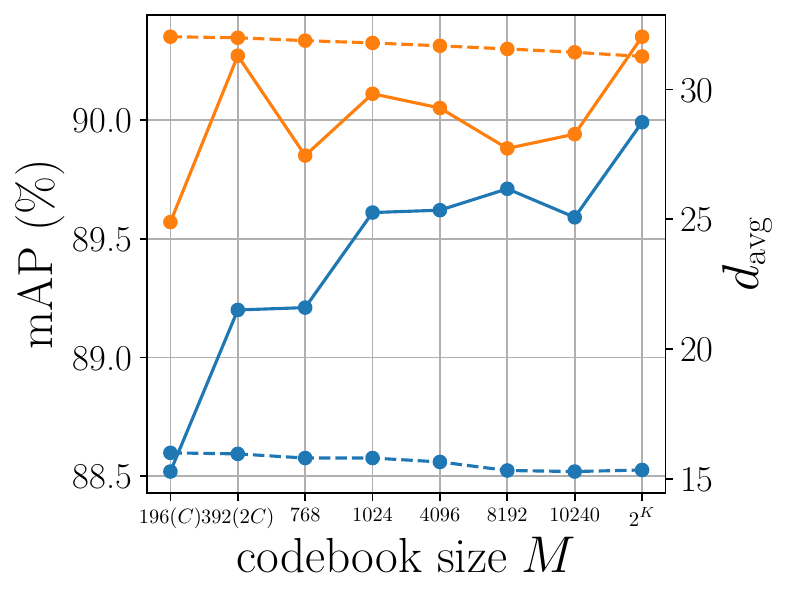}
            \caption{Stanford Cars ($M$)}
        \end{subfigure} &
        \begin{subfigure}[t]{\linewidth}
            \includegraphics[width=\linewidth, height=\textheight, keepaspectratio]{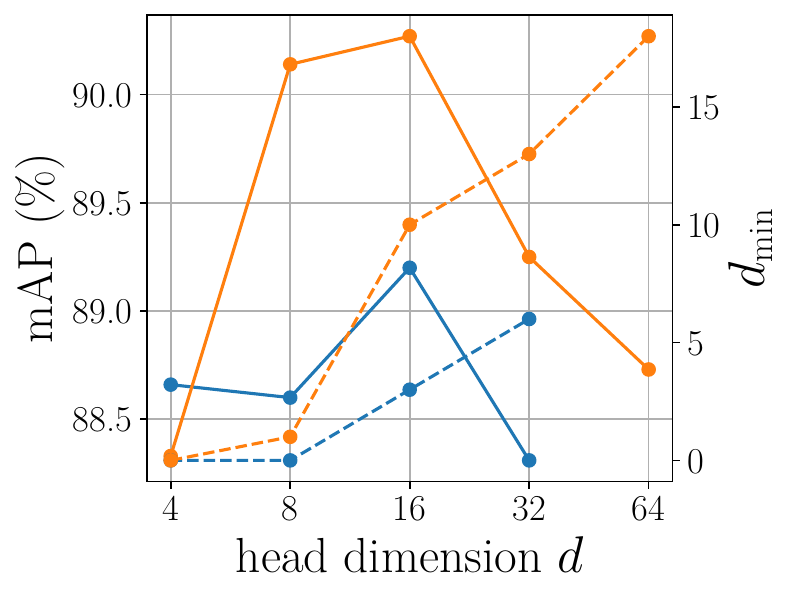}
            \caption{Stanford Cars ($d$)}
        \end{subfigure} \\
        
        \begin{subfigure}[t]{\linewidth}
            \includegraphics[width=\linewidth, height=\textheight, keepaspectratio]{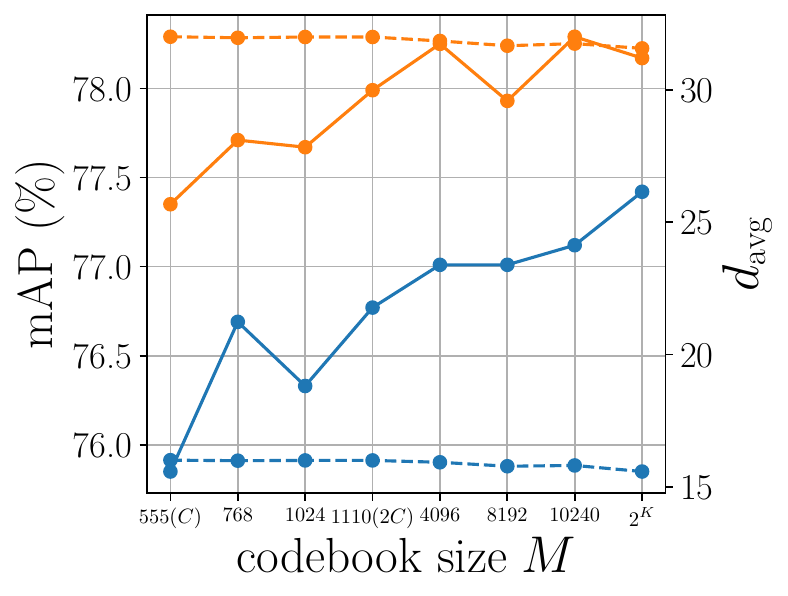}
            \caption{NABirds ($M$)}
        \end{subfigure} &
        \begin{subfigure}[t]{\linewidth}
            \includegraphics[width=\linewidth, height=\textheight, keepaspectratio]{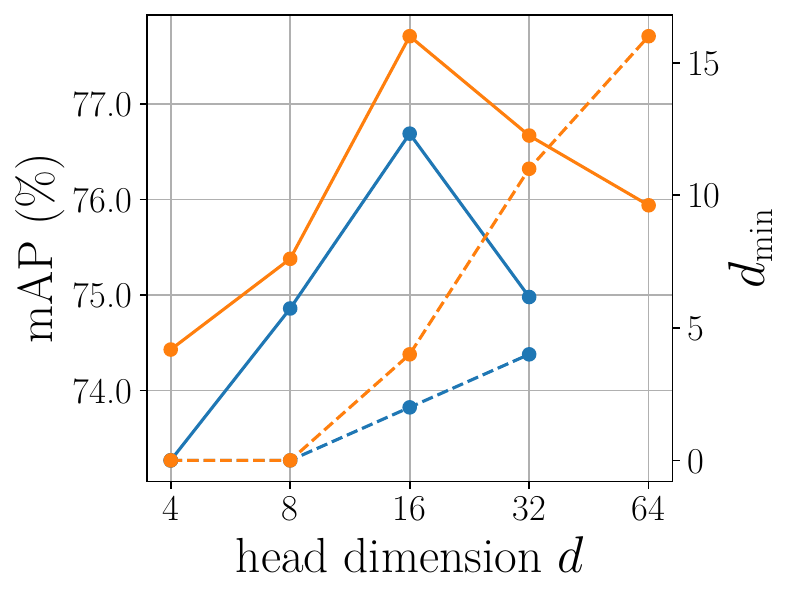}
            \caption{NABirds ($d$)}
        \end{subfigure} \\
        
        \begin{subfigure}[t]{\linewidth}
            \includegraphics[width=\linewidth, height=\textheight, keepaspectratio]{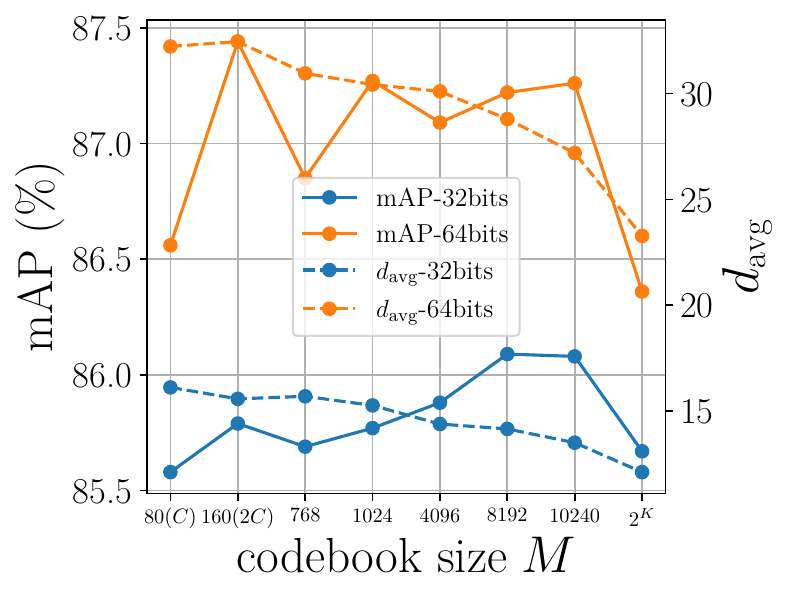}
            \caption{MS COCO ($M$)}
        \end{subfigure} &
        \begin{subfigure}[t]{\linewidth}
            \includegraphics[width=\linewidth, height=\textheight, keepaspectratio]{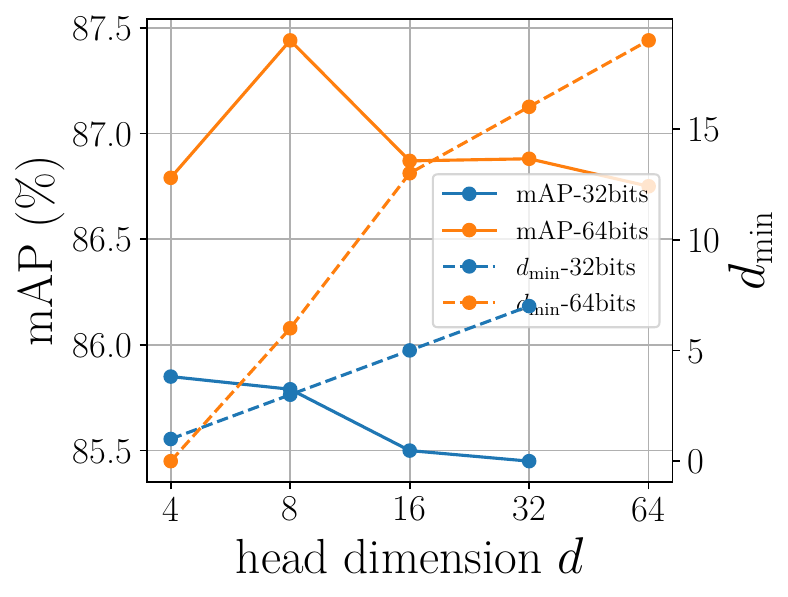}
            \caption{MS COCO ($d$)}
        \end{subfigure}
    \end{tabular}
    \caption{
        mAP w.r.t. codebook size \(M\) (left) and head dimension \(d\) (right) on three benchmark datasets.
    }
    \label{fig_hyparam}
\end{figure}

\subsection{Semantic Quality of Learned Hash Centers} 
To evaluate the semantic expressiveness of our learned hash centers, we design a quantitative framework using CLIP-ViT-L/14’s visual encoder~\citep{Alec-clip-icml-2021} as a reference. For each class $c$, we compute its prototype vector $\mathbf{p}_c$ as a weighted average of sample features: \(
\mathbf{p}_c = \frac{1}{\sum_{\mathbf{x}\in \mathcal{X}_c}\frac{1}{\|\mathbf{y}_\mathbf{x}\|_1}}\sum_{\mathbf{x} \in \mathcal{X}_c} \frac{1}{\|\mathbf{y}_\mathbf{x}\|_1} \text{CLIP}(\mathbf{x}).
\)  
We then construct a reference similarity matrix \( \mathbf{S}^r = (s^r_{ij})_{C \times C} \) with \( s^r_{ij} = \cos(\mathbf{p}_i, \mathbf{p}_j) \), and a corresponding hash center similarity matrix \( \mathbf{S}^h = (s_{ij}^h)_{C \times C} \) based on \( s_{ij}^h = \cos(\mathbf{c}_i, \mathbf{c}_j) \). The alignment between learned hash centers and reference semantics is quantified by the \textbf{Pearson correlation coefficient (PCC)} computed over the strictly upper triangular elements of \( \mathbf{S}^r \) and \( \mathbf{S}^h \). Higher PCC indicates stronger semantic alignment.

As shown in Table~\ref{tab:corr}, untrained centers yield near-zero correlation (row 1). In contrast, our multi-head design achieves average PCCs of $0.310$, $0.271$, and $0.382$ across the three datasets (row 2), validating its ability to learn semantically meaningful hash centers. The single-head variant (row 3) shows notably lower correlations ($0.181$, $0.146$, and $0.131$), highlighting the effectiveness of the multi-head mechanism. Random centers (row 4) exhibit no semantic structure. Among baselines, CSQ, OrthoHash, and MDS—with randomly assigned centers
—show no semantic alignment, while SHC incorporates semantics but still yields low correlations, indicating limited ability to capture fine-grained class relationships.

Finally, we analyze the link between semantic alignment and retrieval. Figure~\ref{fig_mvsc} plots mAP against PCC for three baselines and \textbf{CRH} (64-bit). A clear linear trend emerges—especially on single-label datasets—showing that more expressive hash centers lead to better retrieval. This trend is weaker on MS COCO. 
Notably, CSQ and OrthoHash generate hash centers with nearly identical inter-center distances on this dataset, impeding semantic capture through reassignment and hence limiting performance gains. 
CRH overcomes this through larger randomized codebooks and multi-head design, enhancing diversity and semantic fidelity.

\subsection{Impact of Codebook Size and Head Dimension}
We evaluate how codebook size \(M\) and head dimension \(d\) affect retrieval performance. Figure~\ref{fig_hyparam} shows mAP alongside the minimum (\(d_{\min}\)) and average (\(d_{\text{avg}}\)) distances between learned hash centers under varying configurations at 32/64-bit lengths. 
Note that for \(M = 2^K \) (i.e., $\mathcal{Z}=\{-1,1\}^K$), explicitly constructing $\mathcal{Z}$ is infeasible for large $K$, so we approximate reassignment: for each class \(c\), we set \(\mathbf{z} = \text{sign}(\sum_{\mathbf{x} \in \mathcal{X}_c} \text{sign}(\mathbf{h}_\mathbf{x})/\|\mathbf{y}_\mathbf{x}\|_1)\) as its center if unassigned; otherwise the closest unassigned $\mathbf{z}'\in \mathcal{Z}$ to \(\mathbf{z}\) is assigned.
For \(M\), larger codebooks (vs. \(M = C\)) generally improve mAP but increase computation. On single-label datasets, \(d_{\text{avg}}\) remains near \(K/2\) and declines slowly with \(M\), while on MS COCO it drops significantly, achieving optimal performance at moderate \(M\). We therefore set \(M=2C\) to balance performance and efficiency. For \(d\), best performance occurs when using the smallest power-of-2 dimension satisfying \(d \geq \log_2 M\) (yielding 16, 16, 8 for \(M=2C\) across datasets), maximizing the number of heads while preventing codebook collisions to enhance semantic expressiveness. Conversely, excessively small \(d\) causes collisions that reduce \(d_{\min}\) and degrade performance.

\subsection{Impact of Centers' Update Frequency}\label{section:update-frequency}

Figure~\ref{fig_ui} illustrates how the update interval (i.e., epochs between center updates) affects mAP and PCC (32/64-bit). Longer intervals reduce PCC, indicating weaker semantic alignment, and lead to lower mAP. Although MS~COCO is less sensitive than the single-label datasets, omitting updates ($\infty$) still results in substantial performance drops. These results confirm that more frequent updates generally improve performance.
When training on single-label datasets with ``interval=1'', we observed that roughly 20\% of class centers were updated per step in the first 20 epochs on average, dropping below 1\% afterward, suggesting convergence. Thus, we increased the interval to 5 after 20 epochs, maintaining performance while significantly reducing computation.

\section{Conclusion}
We propose Center-Reassigned Hashing (CRH), an end-to-end deep hashing framework that jointly optimizes hash centers and hash function. By integrating a hash codebook with a multi-head mechanism, CRH dynamically reassigns hash centers to categories, allowing them to adapt to data distributions for enhanced semantic representation—eliminating the need for auxiliary networks or iterative center optimization. Extensive experiments demonstrate that CRH learns semantically discriminative hash centers and achieves state-of-the-art retrieval performance. 

\section{Acknowledgments}
This work is supported in part by National Natural Science Foundation of China (NSFC, Grant No. 62372054, 62006005) and National Key Research and Development Program of China (Grant No. 2022YFC3302200).

\bibliography{aaai2026}

\newpage

\appendix
\section*{Supplementary}
\addcontentsline{toc}{section}{Supplementary}

In the main paper, we propose a supervised image hashing framework, \textbf{CRH}, which dynamically reassigns a hash center for each class from a predefined codebook while jointly optimizing the neural hash function. As this training paradigm operates entirely at the hash code level, it is theoretically applicable to supervised hashing across different modalities, with the potential to deliver strong performance. To validate the generality and effectiveness of our approach beyond image data, we further conduct a series of experiments in the domain of video hashing.

Accordingly, this supplementary material is organized into 2 parts:
\begin{itemize}
    \item additional implementation and experimental details supporting the main paper;

    \item extended evaluations on video hashing to show the modality-generalizability of CRH.
\end{itemize}

\section{Part I. CRH for Image Hashing (supplement)}

\subsection{Notations}
For clarity and ease of reference, Table~\ref{tab:notations} provides a consolidated summary of the key mathematical symbols, variables, and special terms used throughout this paper.

\begin{table}[!htb]
\centering
\resizebox{\columnwidth}{!}{%
\begin{tabular}{c|c|l}
\toprule
\textbf{Category} & \textbf{Symbol} & \textbf{Explanation} \\
\midrule
Dataset & \(\mathcal{X}\) & Dataset containing \(N\) samples\\
& \(N\) & Total number of samples in \(\mathcal{X}\) \\
& \(\mathbf{x}_n\) & Feature vector of the \(n\)-th sample \\
& \(D\) & Dimensionality of feature vectors \\
\midrule
Label & \(\mathbf{y}_n\) & Multi-hot label vector of the \(n\)-th sample \\
& \(C\) & Total number of classes \\
& \(y_{nc}\) & Binary indicator of class \(c\) for the \(n\)-th sample \\
\midrule
Hash Function &\(\text{f}(\cdot)\) & Hash function mapping input to binary-like codes \\
& \(K\) & Length of the hash code \\
& \(\mathbf{h}_n\) & Binary-like code of the $n$-th sample\\
& \(\tanh(\cdot)\) & Hyperbolic tangent activation function \\
& \(\mathbf{v}_n\) & Pre-activation output before $\tanh$ \\
& \(\mathbf{b}_n\) & Final binary hash code \\
& \(\text{sign}(\cdot)\) & Sign function returning \(\{-1, +1\}\) \\
\midrule
Loss Function & \(p_{nc}\) & Softmax probability for class \(c\) on the \(n\)-th sample \\
& \(\text{sim}(\cdot,\cdot)\) & Similarity function \\
& \(margin\) & Margin hyperparameter for similarity calibration \\
& \(s\) & Scaling factor in softmax \\
& \(\|\cdot\|_1\) & \(\ell_1\)-norm (sum of absolute values) \\
& \(\mathbf{1}\) & All-ones vector of appropriate dimension \\
& \(\|\cdot\|_2\) & \(\ell_2\)-norm (Euclidean norm) \\
& \( \text{abs}(\cdot) \) & Element-wise absolute value function \\
& \(\lambda\) & Weight balancing losses \(\mathcal{L}_{\mathrm{CE}}\) and \(\mathcal{L}_{q}\) \\
\midrule
Hash Center & \(\mathcal{Z}\) & Hash codebook\\
Reassignment & \(M\) & Number of candidate centers in \(\mathcal{Z}\) \\
& \(\mathbf{z}_m\) & The \(m\)-th hash center candidate in \(\mathcal{Z}\)\\
& \(\mathbf{c}_c\) & Assigned center for class \(c\) (selected from \(\mathcal{Z}\)) \\
& \(\mathbf{h}_{\mathbf{x}}\) & Continuous hash representation of input \(\mathbf{x}\) \\
& \(\mathcal{X}_c\) & Subset of samples belonging to class \(c\) \\
& \(|\mathcal{X}_c|\) & Number of samples in class \(c\) \\
& \(\mathbf{L}\) & Cost matrix for center reassignment \\
& \(l_{cm}\) & Assignment cost of center \(\mathbf{z}_m\) to class \(c\) \\
& \(j_c^*\) & Index of the optimal center assigned to class \(c\) \\
\midrule
Multi-Head  & \(H\) & Number of heads in multi-head hashing \\ Mechanism
& \(\mathbf{z}^h_m\) & Sub-vector of \(\mathbf{z}_{m}\) for head \(h\) (size \(d=K/H\)) \\
& \(d\) & Dimensionality of each sub-vector \\
& \(\mathcal{Z}^h\) & Codebook for head $h$ \\
& \(\mathbf{h}^h_\mathbf{x}\) & Sub-vector of \(\mathbf{h}_{\mathbf{x}}\) in head \(h\) \\
& \(\mathbf{L}^h\) & Cost matrix for head $h$ \\
& \(\mathbf{c}^h_c\) & Center assigned to class \(c\) in head \(h\) \\
\bottomrule
\end{tabular}
}
\caption{Summary of notations used in the CRH paper.}
\label{tab:notations}
\end{table}

\subsection{Implementation Details}
In the main paper, the cost matrix $\mathbf{L} = (l_{cm})_{C \times M}$ is computed as:
\begin{equation}\label{eq:lcm}
    l_{cm} = \frac{1}{\sum_{\mathbf{x} \in \mathcal{X}_c} \frac{1}{\|\mathbf{y}_\mathbf{x}\|_1}} \sum_{\mathbf{x} \in \mathcal{X}_c} \frac{1}{\|\mathbf{y}_\mathbf{x}\|_1} \left\| \text{sign}(\mathbf{h}_\mathbf{x}) - \mathbf{z}_m \right\|^2_2.
\end{equation}
Note that computing the cost matrix $\mathbf{L}$ for hash center updates requires the outputs \( \mathbf{h_x} \) of all training data through network f and involves a summation over data points as in Eq.\ref{eq:lcm}. To avoid the computationally expensive reprocessing of the entire training set, we directly utilize the network's outputs to incrementally calculate $\mathbf{L}$ during the previous epoch of training the hash function.

\subsection{Additional Results} 
We conducted evaluations on two additional datasets:  

\textbf{NUS-WIDE}~\citep{Tat-nuswide-civr-2009}: a multi-label dataset with 81 categories. Following~\citep{Han-DHN-aaai-2016}, images containing the 21 most frequent tags are selected. The database, training, and query partitions contain 149,736, 10,500, and 2,100 images, respectively.  

\textbf{ImageNet}~\citep{Jia-imagenet-cvpr-2009}: a single-label dataset consisting of 1,000 classes. Following~\citep{Zhangjie-hashnet-iccv-2017}, a subset of 100 classes is selected. The training and validation sets of these classes are used as the database (128,503 images) and query set (5,000 images), respectively. From the database, 130 images per class (13,000 in total) are chosen for model training.  

The mAP@5,000 for NUS-WIDE and mAP@1,000 for ImageNet are reported in Table~\ref{tab:supp_results}.  

\begin{table}[t]  
\centering   
\resizebox{\columnwidth}{!}{
\begin{tabular}{l|ccc|ccc}
        \Xhline{0.8pt}
        \multirow{2}{*}{Methods} & \multicolumn{3}{c|}{\textbf{NUS-WIDE} (mAP@5K)} & \multicolumn{3}{c}{\textbf{ImageNet} (mAP@1K)}  \\
        \cline{2-7}
           & 16          & 32         & 64         & 16           & 32          & 64          \\
           \hline  
DTSH       & \underline{80.9}        & 82.4       & 83.1       & 72.1         & 77.0        & 78.5        \\
HashNet    & 79.9        & 81.9       & 82.9       & 54.2         & 71.0        & 76.5        \\
GreedyHash & 78.2        & 81.7       & 82.7       & 76.7         & 80.0        & 80.2        \\
CSQ        & 79.0        & 81.5       & 82.3       & 77.3         & 79.6        & 80.5        \\
OrthoHash  & \textbf{81.0}        & \underline{82.7}       & \underline{83.6}       & \underline{80.9}         & \underline{84.2}        & \underline{85.8}        \\
MDS        & 76.6        & 79.9       & 80.7       & 76.4         & 80.5        & 81.2        \\
SHC        & 75.4        & 79.5       & 80.4       & 79.1         & 80.6        & 81.1        \\
           \hline
CRH (Ours)        & \textbf{81.0}        & \textbf{82.9}       & \textbf{84.1}       & \textbf{84.2}         & \textbf{86.0}        & \textbf{87.2}    \\
\Xhline{0.8pt}
\end{tabular}
}
\caption{Comparison of retrieval performance (mAP, \%) between our method and deep hashing baselines on NUS-WIDE and ImageNet datasets at multiple bit lengths.}  
\label{tab:supp_results} 
\end{table}

\subsection{Impact of hyperparameter $\lambda$}
\begin{figure*}[t]
    \centering
    \begin{subfigure}{0.28\textwidth}
        \includegraphics[width=\linewidth]{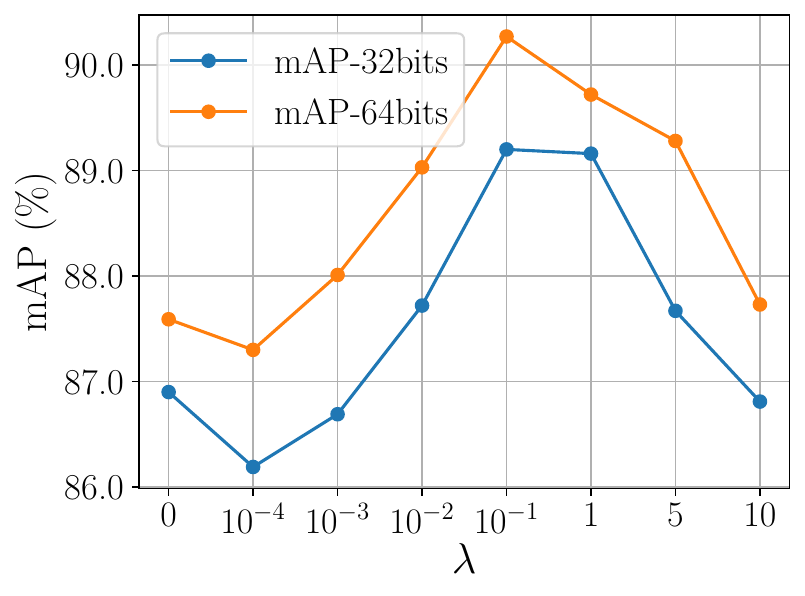}
        \caption{Stanford Cars}
        \label{subfig:cars_lmd}
    \end{subfigure}
    \hspace{0.04\textwidth}%
    \begin{subfigure}{0.28\textwidth}
        \includegraphics[width=\linewidth]{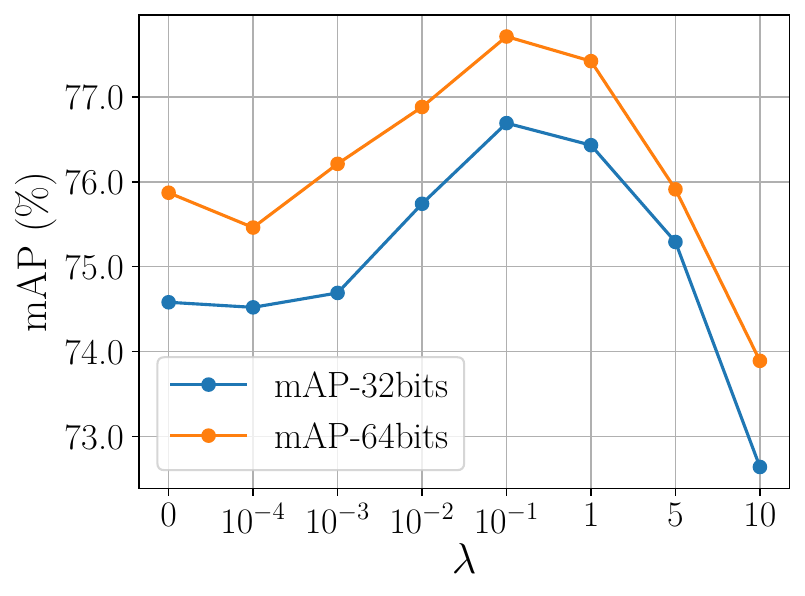}
        \caption{NABirds}
        \label{subfig:birds_lmd}
    \end{subfigure}
    \hspace{0.04\textwidth}%
    \begin{subfigure}{0.28\textwidth}
        \includegraphics[width=\linewidth]{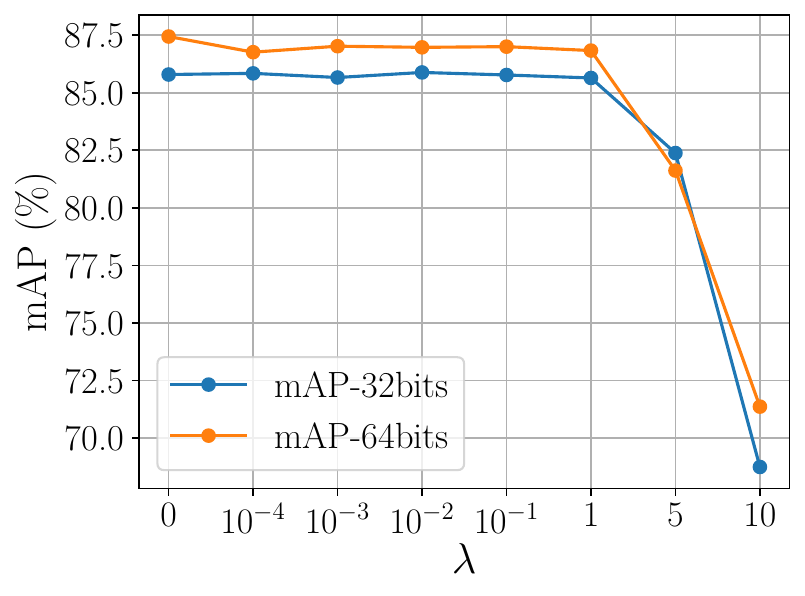}
        \caption{MS COCO}
        \label{subfig:coco_lmd}
    \end{subfigure}
    
    \caption{mAP scores (\%) vs. the hyperparameter $\lambda$ values under 32/64-bit configurations across three datasets.}
    \label{fig_lmd}
\end{figure*}

Figure~\ref{fig_lmd} shows the impact of the quantization loss weight $\lambda$ on retrieval performance. For single-label datasets, incorporating the quantization loss $\mathcal{L}_q$ is clearly advantageous. In particular, setting $\lambda = 0.1$ achieves the highest performance, significantly outperforming the baseline without quantization regularization (i.e., $\lambda = 0$). In contrast, for the multi-label dataset, introducing $\mathcal{L}_q$ offers no notable benefit and can even impair performance. This occurs because the cross-entropy loss $\mathcal{L}_\mathrm{CE}$ drives the model's output for each sample toward a mean of its associated class centers, which is not strictly binarized, making additional quantization regularization non-beneficial. We therefore disable the quantization loss in the multi-label setting by setting $\lambda = 0$ throughout all relevant experiments.

\subsection{Training and Inference Time Analysis}
\textbf{Training time.} Table~\ref{tab:time} reports the training time of hash-center-based methods across three datasets under the 64-bit setting. Overall, the average training time per epoch is comparable among methods. CRH incurs a slightly higher per-epoch cost—exceeding others by only 0.6\%, 0.2\%, and 0.6\% on the three datasets, respectively—primarily due to the additional computation of the cost matrix $\mathbf{L}$ during training (see implementation details). However, the time required for a single center reassignment in CRH is less than 50 milliseconds, rendering it negligible in practice.

In contrast, MDS and SHC involve a separate stage for hash center generation, which introduces substantial overhead. This cost becomes particularly significant on datasets with a large number of classes, where center generation can even surpass the time required for hash function training. Despite its dynamic reassignment mechanism, CRH achieves total training times comparable to CSQ and OrthoHash, highlighting its overall training efficiency.

\textbf{Inference time.} All methods adopt the same ResNet-34 backbone for feature extraction, resulting in nearly identical inference times for encoding individual samples into hash codes.

\begin{table}[ht!]
    \centering
    \resizebox{\columnwidth}{!}{
        \begin{tabular}{llccc}
        \Xhline{0.8pt}
            Method                     & Phase  & \textbf{Stanford cars} & \textbf{NABirds}  & \textbf{MS COCO} \\
            \hline
\multirow{2}{*}{CSQ}       & Train  & 10.085s       & 33.930s  & 10.144s \\
                           & Total  & 50.426m       & 169.649m & 5.072m  \\
                           \hline
\multirow{2}{*}{OrthoHash} & Train  & 9.926s        & 33.718s  & 10.064s \\
                           & Total  & 49.631m       & 168.591m & 5.032m  \\
                           \hline
\multirow{3}{*}{MDS}       & Gen    & 33.670m       & 255.884m & 0.753m  \\
                           & Train  & 10.278s       & 34.109s  & 10.331s \\
                           & Total  & 85.060m       & 426.429m & 5.919m  \\
                           \hline
\multirow{3}{*}{SHC}       & Gen    & 65.311m       & 424.762m & 11.191m \\
                           & Train  & 9.899s        & 33.727s  & 10.000s \\
                           & Total  & 114.804m      & 593.396m & 16.191m \\

                           \Xhline{0.6pt}
\multirow{3}{*}{CRH}       & Update & 0.019s        & 0.045s   & 0.010s  \\
                           & Train  & 10.111s       & 33.938s  & 10.199s \\
                           & Total  & 50.581m       & 169.748m & 5.103m \\
                           \Xhline{0.8pt}
        \end{tabular}
    }
    \caption{
    \textbf{Training Time (in seconds ``s'' and minutes “m”) for different methods at 64-bit setting across three datasets, measured on an NVIDIA A800 GPU (80GB).}
    Note that ``Train'' denotes the average training time per epoch. For MDS and SHC, ``Gen'' refers to the one-time cost of hash center generation. For CRH, ``Update'' indicates the average time for a single center reassignment. ``Total'' summarizes the overall training time, including ``Train'', ``Gen'', or ``Update'' as applicable.}
    \label{tab:time}
\end{table}

\section{Part II. CRH for Video Hashing (newly added)}

It's noteworthy that CRH, while effective for image modality, can also be extended to other modalities such as video. In this part, we further testify the effectiveness and versatility of CRH by applying it to video hashing, thereby validating its superior performance across different modalities.

\subsection{Methodologies}

The CRH approach can be readily extended to video hashing with minimal modification. \textbf{\textit{The only notable change}} involves replacing the single-image input with sequential video frames and substituting the image hash function with its video counterpart. All other components and procedures remain intact, demonstrating the seamless adaptability of CRH across different modalities.

\subsection{Experimental Setups}

\subsubsection{Datasets}
We evaluate our training strategy performance on video hashing with two widely adopted single-label datasets: \textbf{ActivityNet} and \textbf{FCVID}.

\begin{itemize}
    \item \textbf{ActivityNet}~\citep{Fabian-ActivityNet-CVPR-2015} is a single-label dataset with 200 human activity categories, comprising 12,535 training and 2,413 test videos. We use the training set for the retrieval database, and the test set serving as queries. 

    \item  \textbf{FCVID}~\citep{Yu-Exploiting-IEEE-2018} is a single-label comprehensive dataset featuring 239 categories with a wide range of topics, and splited into 66,694 training and 22,204 test images. Consistent with ActivityNet, we adopt the training set for retrieval, and the test set for queries. 
\end{itemize}

\begin{table*}[ht!]\small
    \centering
    \resizebox{\textwidth}{!}{
    \begin{tabular}{l|ccc|ccc|ccc|ccc}
        \Xhline{0.8pt}
        \multirow{3}{*}{Methods} & \multicolumn{6}{c|}{\textbf{ActivityNet} (mAP@all)} & \multicolumn{6}{c}{\textbf{FCVID} (mAP@all)} \\
        \cline{2-13}
        & \multicolumn{3}{c|}{\textbf{CRH's Training Strategy}} & \multicolumn{3}{c|}{\textbf{Original}} & \multicolumn{3}{c|}{\textbf{CRH's Training Strategy}} & \multicolumn{3}{c}{\textbf{Original}}\\
        \cline{2-13}
        & 16 bits & 32 bits & 64 bits & 16 bits & 32 bits & 64 bits & 16 bits & 32 bits & 64 bits & 16 bits & 32 bits & 64 bits\\
        \hline       
        AVHash (w/o audio) & 91.18 & 92.08 & 92.59 & 80.72 & 85.01 & 86.89 & 92.45 & 93.33 & 93.46 & 91.78 & 92.47 & 92.89\\
        AVH & 41.73 & 46.02 & 48.76 & 15.86 & 31.68 & 46.83 & 58.79 & 64.08 & 68.32 & 15.05 & 25.34 & 43.59\\
        SRH & 19.67 & 30.34 & 39.40 & 16.65 & 29.85 & 37.22 & 32.28 & 43.62 & 51.95 & 20.01 & 32.54 & 44.70\\
        \Xhline{0.8pt}
    \end{tabular}
    }    
    \caption{Comparison of retrieval performance (mAP, \%) between CRH's and video hashing baselines' training strategy on two datasets at multiple bit lengths.}
    \label{tab:mAP_video}
\end{table*}

\begin{table*}[ht!]\small
    \centering
    \setlength{\tabcolsep}{23pt}
    \begin{tabular}{l|ccc|ccc}
        \Xhline{0.8pt}
        \multirow{2}{*}{Methods} & \multicolumn{3}{c|}{\textbf{ActivityNet} (mAP@all)} & \multicolumn{3}{c}{\textbf{FCVID} (mAP@all)} \\
        \cline{2-7}
        & 16 bits & 32 bits & 64 bits & 16 bits & 32 bits & 64 bits \\
        \hline       
        CRH   & \textbf{91.18} & \textbf{92.08} & \textbf{92.59} & \textbf{92.45} & \textbf{93.33} & \textbf{93.46} \\
        CRH-M & \underline{90.91} & \underline{91.58} & 91.45 & \underline{92.04} & \underline{92.62} & \underline{93.17} \\
        CRH-U & 89.68 & 91.24 & \underline{91.74} & 90.35 & 91.56 & 92.33 \\
        \Xhline{0.8pt}
    \end{tabular}
    \caption{Comparison of retrieval performance (mAP, \%) between CRH and its ablated variants (CRH-M, CRH-U) using AVHash (w/o audio) as model.}
    \label{tab:ACRH_video}
\end{table*}

\subsubsection{Metrics and Baselines}
We evaluate retrieval performance using mean Average Precision (mAP@all) for both ActivityNet and FCVID. To further demonstrate the effectiveness of CRH's training strategy on the video modality, we adapt three state-of-the-art supervised video hashing methods by replacing their original training components with the CRH's framework, and compare their performance before and after this replacement:
\begin{itemize}
    \item \textbf{SRH}~\cite{SRH-Yun-ACMM-2016}: An LSTM-based approach that models video structures while preserving pairwise similarity relationships in the binary embedding space.

    \item \textbf{AVH}~\cite{AVH-Yingxin-TransCogn-2021}: Attention-based Video Hashing that combines CNN and LSTM backbones to learn compact hash codes by capturing cross-frame structural dependencies.

    \item \textbf{AVHash}~\cite{AVHash-Yuxiang-ACMM-2024}: A Transformer-based architecture employing cross-attention to fuse visual and audio modalities, trained with a supervised triplet loss to produce semantically enriched multimodal hash codes.
\end{itemize}
It is worth noting that for AVHash, we retain only the visual branch and discard the audio branch to ensure a fair comparison and to enable the use of CRH's training procedure.

\subsubsection{Implementation Details}
When trained with CRH's strategy, all baseline models are optimized using Adam with coefficients $(\beta_1, \beta_2) = (0.5, 0.999)$ and a weight decay of $10^{-5}$. The margin $margin$ is set to $0.2$ following~\citep{Jiun-orthoHash-nips-2021}, and the scale parameter $s$ is computed as $\sqrt{2}\log(C-1)$ according to~\citep{Xiao-adacos-cvpr-2019}. The hash codebook size $M$ is fixed at $2C$. Training uses a cosine annealing schedule with an initial learning rate of $10^{-4}$. The hyperparameter $\lambda$ is set to $0.1$ for both ActivityNet and FCVID. The head dimension $d$ is configured as $16$ for both datasets, and centers are updated by default using the greedy algorithm. We train with a batch size of 128 for 300 epochs.

\subsection{Results}
The mAP results are summarized in Table~\ref{tab:mAP_video}, where ``Original'' denotes each baseline trained with its own strategy. As shown in each row, after adopting CRH's training scheme, all baseline models consistently achieve substantially better performance across both datasets, with improvements ranging from 4\% to 62\% on ActivityNet and 0.6\% to 74.4\% on FCVID. 

For ActivityNet, the most substantial improvements appear at shorter code lengths (16 and 32 bits), with diminishing returns as the code length increases. In contrast, on FCVID, CRH provides robust performance gains across all hash code lengths. Notably, AVH and SRH perform significantly worse on both datasets, which is likely because their optimization objectives rely on the cross-entropy between hash codes and labels, making them less effective at capturing the inter- and intra-class relationships inherent in video features.

\begin{table*}[ht!]\small
    \centering
    \setlength{\tabcolsep}{22pt}
        \begin{tabular}{l|ccc|ccc}
            \Xhline{0.8pt}
            \multirow{2}{*}{Methods} & \multicolumn{3}{c|}{\textbf{ActivityNet} (mAP@all)} & \multicolumn{3}{c}{\textbf{FCVID} (mAP@all)} \\
            \cline{2-7}
            & 16 bits & 32 bits & 64 bits & 16 bits & 32 bits & 64 bits \\
            \hline       
            CSQ  & 87.36  & 89.86  & 91.59  & 91.15  & 92.02  & 92.87 \\
            CSQ$_U$  & \textbf{88.48} & \textbf{90.64} & \textbf{91.80} & \textbf{91.52} & \textbf{92.79} & \textbf{92.95} \\
            \hline 
            MDS  & 81.76  & 88.78  & 89.91  & 90.19  & 90.30  & 91.56 \\
            MDS$_U$  & \textbf{84.01} & \textbf{89.37} & \textbf{90.87} & \textbf{90.44} & \textbf{91.37} & \textbf{92.68} \\
            \hline 
            OrthoHash  & 86.92  & \textbf{90.44}  & \textbf{90.64}  & 90.25  & 90.49  & 91.39  \\
            OrthoHash$_U$ & \textbf{87.59} & \textbf{90.44} & \textbf{90.64} & \textbf{90.59}  & \textbf{91.04}  & \textbf{91.82} \\       
            \Xhline{0.8pt}
        \end{tabular}
    \caption{Performance comparison (mAP, \%) of hash-center-based training methods (using AVHash (w/o audio) as model) with/without the proposed update mechanism.} 
    \label{tab:Absl_video}
\end{table*}

\begin{table*}[ht!]\small
    \centering
    \setlength{\tabcolsep}{15pt}
        \begin{tabular}{l|ccc|ccc}
            \Xhline{0.8pt}
            \multirow{2}{*}{Methods} & \multicolumn{3}{c|}{\textbf{ActivityNet} (mAP@all)} & \multicolumn{3}{c}{\textbf{FCVID} (mAP@all)} \\
            \cline{2-7}
            & 16 bits & 32 bits & 64 bits & 16 bits & 32 bits & 64 bits \\
            \hline       
            Seed   & 90.83 $\scriptstyle{\pm\,0.35}$ & 92.21 $\scriptstyle{\pm\,0.17}$ & 92.66 $\scriptstyle{\pm\,0.07}$ & 92.32 $\scriptstyle{\pm\,0.18}$ & 93.34 $\scriptstyle{\pm\,0.06}$ & 93.45 $\scriptstyle{\pm\,0.04}$\\
            Init   & 91.13 $\scriptstyle{\pm\,0.09}$ & 92.29 $\scriptstyle{\pm\,0.40}$ & 92.60 $\scriptstyle{\pm\,0.14}$ & 92.49 $\scriptstyle{\pm\,0.14}$ & 93.28 $\scriptstyle{\pm\,0.06}$ & 93.48 $\scriptstyle{\pm\,0.03}$\\
            Init-H & 90.79 $\scriptstyle{\pm\,0.39}$   & 92.07 $\scriptstyle{\pm\,0.07}$ & 92.58 $\scriptstyle{\pm\,0.04}$ & 92.00 $\scriptstyle{\pm\,0.07}$ & 92.57 $\scriptstyle{\pm\,0.25}$ & 93.06 $\scriptstyle{\pm\,0.11}$\\     
            \Xhline{0.8pt}
        \end{tabular}
    \caption{Performance impact of initialization randomness and update algorithms (mean±std mAP over 3 runs). ``Seed'': random update mechanism; ``Init'': varied center initializations; ``Init-H'': ``Init'' with Hungarian-algorithm updates.}
    \label{tab:random_video}
\end{table*}

\begin{figure*}[!htp]
    \centering
    \subfloat[ActivityNet ($M$)]{%
        \includegraphics[width=0.245\textwidth]{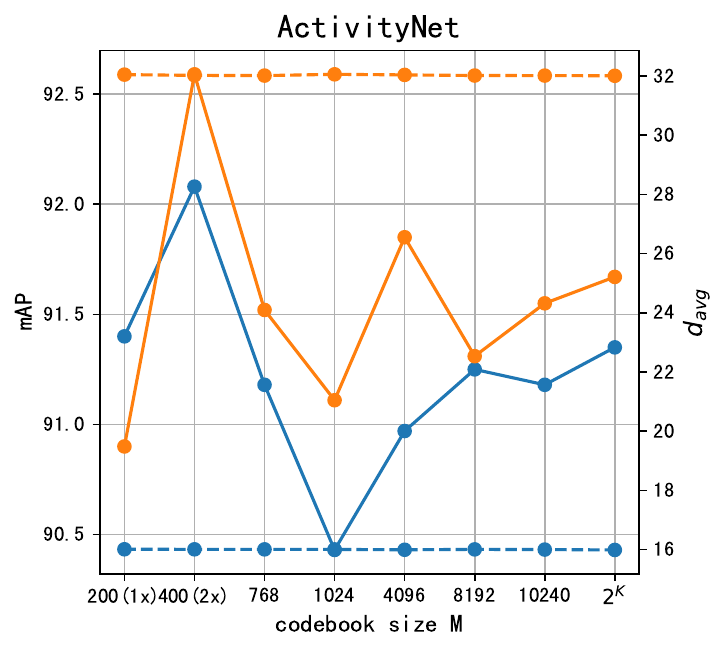}
    }
    \subfloat[FCVID ($M$)]{%
        \includegraphics[width=0.245\textwidth]{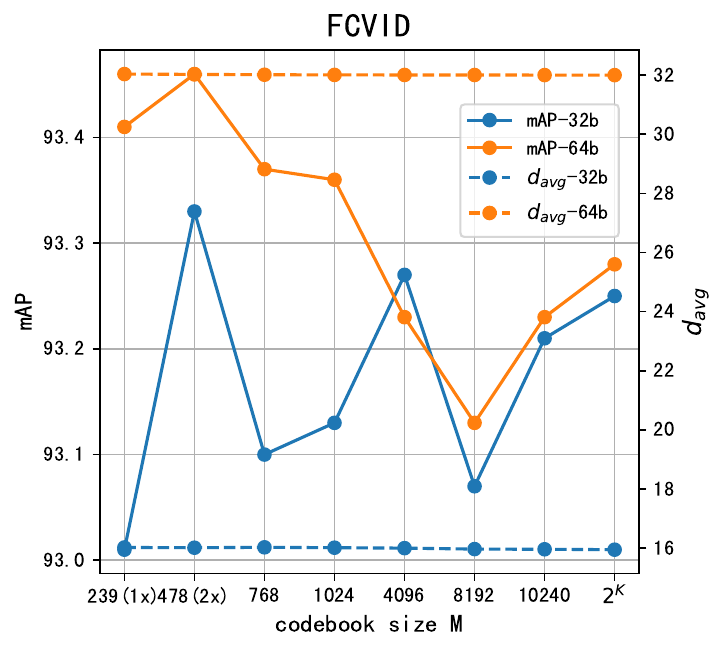}
    }
    \subfloat[ActivityNet ($d$)]{%
        \includegraphics[width=0.245\textwidth]{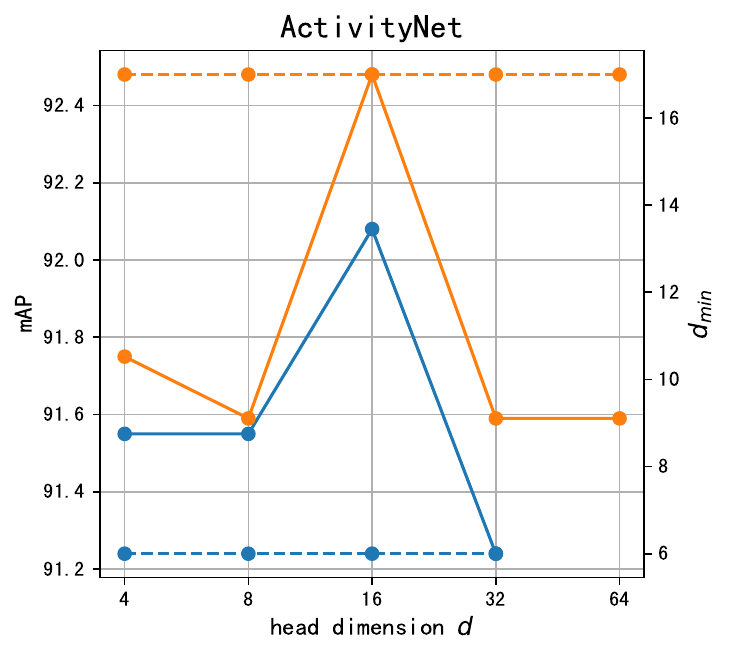}
    }
    \subfloat[FCVID ($d$)]{%
        \includegraphics[width=0.245\textwidth]{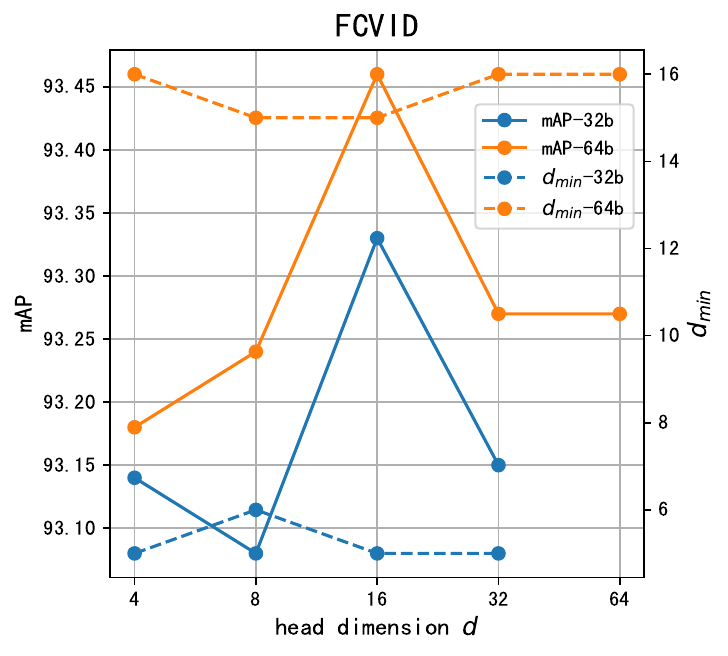}
    }

    \caption{mAP w.r.t. codebook size \(M\) (top) and head dimension \(d\) (bottom), evaluated using 32-bit and 64-bit hash codes on two benchmark datasets with AVHash (w/o audio) as model.}
    \label{fig_hyparam_video}
\end{figure*}

\subsection{Ablation Studies}

To evaluate the contributions of the two key components of CRH's training strategy for video hashing—namely, the center reassignment mechanism and the multi-head architecture—we conduct systematic ablation studies by combining the AVHash model (selected due to its superior performance in Table~\ref{tab:mAP_video}) with CRH's training framework. Specifically, we investigate two variants:
\begin{itemize}
     \item \textbf{CRH-U}: Disables center reassignment and consequently the multi-head mechanism, resulting in a fixed-center approach similar to CSQ/OrthoHash; 

    \item \textbf{CRH-M}: Retains reassignment but removes the multi-head mechanism (i.e., uses a single head, $H = 1$).
\end{itemize}

As shown in Table~\ref{tab:ACRH_video}, CRH-M consistently outperforms CRH-U across all settings except for 64-bit codes on ActivityNet, achieving average relative mAP gains of 0.8\% and 1.31\% on ActivityNet and FCVID, respectively. This demonstrates that the dynamic center reassignment mechanism effectively adapts to data distributions, yielding improved semantic representations for the video modality, consistent with its impact on image data. The complete CRH model further improves upon CRH-M by leveraging the multi-head architecture, attaining additional average gains of 0.67\% and 0.47\%, which confirms that multiple heads enable finer hash center refinement for video hashing.

To further assess generalizability within the video hashing domain, we integrate CRH’s update mechanism into three existing center-based training methods (CSQ, MDS, and OrthoHash), denoted as $(\cdot)_U$, while uniformly employing the visual branch of AVHash as the backbone. These enhanced variants maintain their original codebook structures (size $C$) without multi-head extensions. As reported in Table~\ref{tab:Absl_video}, they achieve consistent performance improvements across almost all configurations, demonstrating the broad applicability of our approach to enhancing center-based hashing methods, even for video inputs.

\subsection{Robustness to Randomness}  
Our algorithm introduces stochasticity through two components: (1) random initialization of hash centers and (2) the greedy center update procedure. To assess their impact on video hashing, we conduct controlled experiments using the AVHash model:
\begin{itemize}
    \item \textbf{Initialization Robustness}: Three trials with different random initial center assignments while keeping all other parameters fixed (``Init'').
    
    \item \textbf{Update Procedure Robustness}: Three runs with varying random seeds but identical initial centers (``Seed'').
\end{itemize}
As reported in Table~\ref{tab:random_video}, the consistently low standard deviations across all settings indicate that CRH maintains strong stability against these stochastic factors in the video modality.

We further compare our greedy algorithm (``Init'')  with the Hungarian algorithm (``Init-H''). While both approaches achieve similar mAP results, the greedy algorithm consistently offers notable advantages: (1) it yields slightly higher mAP (with average relative improvements of 0.17\% on ActivityNet and 0.53\% on FCVID), attributed to the stochasticity from its random class order which enables broader exploration of the solution space; and (2) it operates with substantially lower computational complexity ($\mathcal{O}(HCM)$ vs. $\mathcal{O}(HC^2M)$).

\subsection{Impact of Codebook Size and Head Dimension}
Using the AVHash model, we evaluate how the codebook size \(M\) and head dimension \(d\) influence video hashing retrieval performance. Figure~\ref{fig_hyparam_video} reports mAP along with the minimum (\(d_{\min}\)) and average (\(d_{\text{avg}}\)) distances between learned hash centers under various configurations at 32- and 64-bit code lengths. 

For the codebook size \(M\), larger codebooks (compared to \(M = 2C\)) generally lead to lower mAP and increased computational costs. Additionally, on both datasets, \(d_{\text{avg}}\) stays close to \(K/2\) and decreases slowly as \(M\) grows, indicating a higher likelihood of semantic confusion between categories when \(M\) is large. Thus, we set \(M=2C\) to balance performance and efficiency.

Regarding the head dimension \(d\), the best performance is also achieved when using the smallest power-of-two dimension satisfying \(d \geq \log_2 M\) (yielding \(d=16\) for \(M=2C\) on both datasets), consistent with observations in image hashing. This choice maximizes the number of heads while preventing codebook collisions, thereby enhancing semantic expressiveness. However, unlike image hashing, the relationship between \(d\) and \(d_{\min}\) is less pronounced for video data. This can be mainly attributed to the inherent discriminability of spatiotemporal features in video data and the enhanced representational capacity of sequential models like Transformers. These architectures autonomously capture fine-grained motion patterns and temporal dependencies, maintaining high inter-center distances without relying on multi-head mechanism.

\end{document}